\documentclass[letterpaper]{article} % DO NOT CHANGE THIS
\usepackage{aaai2026}  % DO NOT CHANGE THIS
\usepackage{times}  % DO NOT CHANGE THIS
\usepackage{helvet}  % DO NOT CHANGE THIS
\usepackage{courier}  % DO NOT CHANGE THIS
\usepackage[hyphens]{url}  % DO NOT CHANGE THIS
\usepackage{graphicx} % DO NOT CHANGE THIS
\usepackage{natbib}  % DO NOT CHANGE THIS AND DO NOT ADD ANY OPTIONS TO IT
\usepackage{caption} % DO NOT CHANGE THIS AND DO NOT ADD ANY OPTIONS TO IT
\usepackage{algorithm}
\usepackage{algorithmic}

\usepackage[cmyk]{xcolor}% 
\definecolor{my_blue}{cmyk}{0.04, 0.02, 0, 0}
\definecolor{cvpr_blue}{cmyk}{0.72,0.34,0,0.26}
\definecolor{my_green}{cmyk}{0.04, 0, 0.06, 0.02}
\definecolor{acl_green}{cmyk}{0.74,0,0.21,0.40}
\definecolor{mycyan}{cmyk}{0.065, 0, 0, 0}

\usepackage{pifont} % 
\usepackage{colortbl}
\usepackage[most]{tcolorbox}

\usepackage{algorithm}
\usepackage{algorithmic}
\usepackage{multirow}
\usepackage{booktabs}

\usepackage[cmyk]{xcolor}% 
\definecolor{my_blue}{cmyk}{0.04, 0.02, 0, 0}
\definecolor{cvpr_blue}{cmyk}{0.72,0.34,0,0.26}
\definecolor{my_green}{cmyk}{0.04, 0, 0.06, 0.02}
\definecolor{acl_green}{cmyk}{0.74,0,0.21,0.40}
\definecolor{mycyan}{cmyk}{0.065, 0, 0, 0}

\usepackage{pifont} % 
\usepackage{colortbl}
\usepackage[most]{tcolorbox}

\usepackage{algorithm}
\usepackage{algorithmic}
\usepackage{multirow}
\usepackage{booktabs}
\definecolor{cvprblue}{rgb}{0.21,0.49,0.74}

\usepackage{newfloat}
\usepackage{listings}
\DeclareCaptionStyle{ruled}{labelfont=normalfont,labelsep=colon,strut=off} % DO NOT CHANGE THIS
\floatstyle{ruled}
\newfloat{listing}{tb}{lst}{}
\floatname{listing}{Listing}
\title{CCFQA: A Benchmark for Cross-Lingual and Cross-Modal \\ Speech and Text Factuality Evaluation}

\author{
Yexing Du$^{1,2}$, Kaiyuan Liu$^{1,2}$, Youcheng Pan$^{2}$, Zheng Chu$^{1}$,\\ Bo Yang$^{2}$, Xiaocheng Feng$^{1}$, Ming Liu\thanks{ Corresponding author.}$^{1,2}$, Yang Xiang$^{*}$$^{2}$  \\
}

\affiliations{
\textsuperscript{\rm 1}Harbin Institute of Technology\\ 
\textsuperscript{\rm 2}Pengcheng Laboratory \\
\{yxdu, mliu\}@ir.hit.edu.cn, \{panych, xiangy\}@pcl.ac.cn
}

\usepackage{bibentry}
\begin{document}

\maketitle

\begin{abstract}
As Large Language Models (LLMs) are increasingly popularized in the multilingual world, ensuring hallucination-free factuality becomes markedly crucial. 
However, existing benchmarks for evaluating the reliability of Multimodal Large Language Models (MLLMs) predominantly focus on textual or visual modalities with a primary emphasis on English, which creates a gap in evaluation when processing multilingual input, especially in speech.
To bridge this gap, we propose a novel \textbf{C}ross-lingual and \textbf{C}ross-modal \textbf{F}actuality benchmark (\textbf{CCFQA}). 
Specifically, the CCFQA benchmark contains parallel speech-text factual questions across 8 languages, designed to systematically evaluate MLLMs' cross-lingual and cross-modal factuality capabilities. 
Our experimental results demonstrate that current MLLMs still face substantial challenges on the CCFQA benchmark. 
Furthermore, we propose a few-shot transfer learning strategy that effectively transfers the Question Answering (QA) capabilities of LLMs in English to multilingual Spoken Question Answering (SQA) tasks, achieving competitive performance with GPT-4o-mini-Audio using just 5-shot training.
We release CCFQA as a foundational research resource to promote the development of MLLMs with more robust and reliable speech understanding capabilities.

\end{abstract}

\begin{links}
\link{Dataset}{https://github.com/yxduir/ccfqa}
 \end{links}

\section{Introduction}
\begin{figure}[t]
\includegraphics[width=\columnwidth]{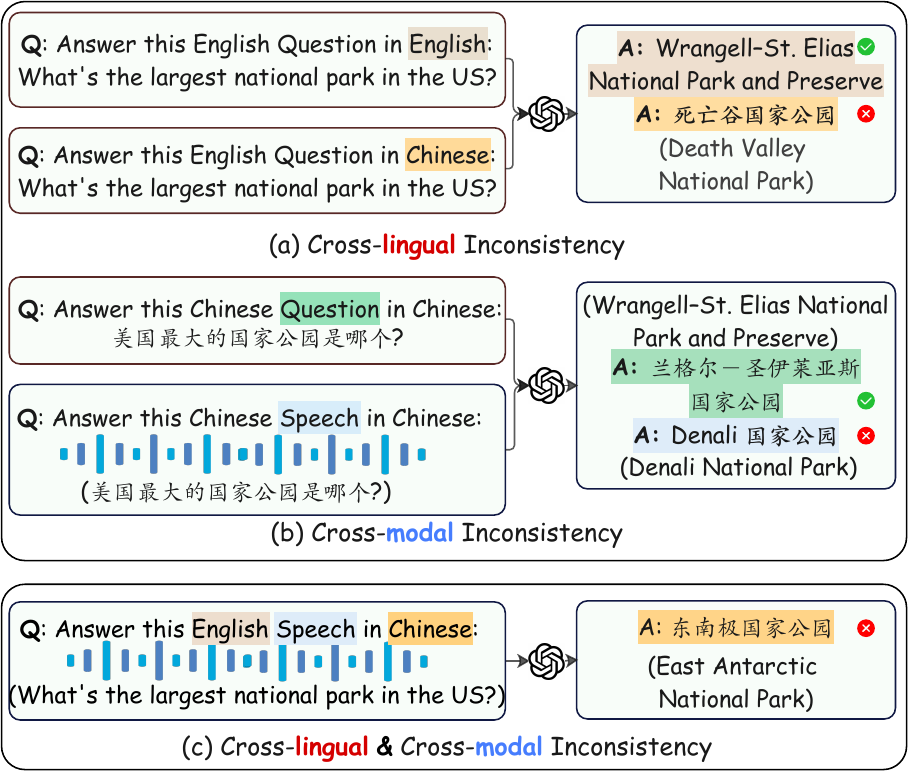}
\caption{Factual Inconsistency in MLLMs. (a) Cross-lingual Inconsistency: inconsistent answers for the questions across different languages;
(b) Cross-modal Inconsistency: inconsistent answers for the questions across different modalities;
(c) Cross-lingual \& Cross-modal Inconsistency.}
\label{intro}
\end{figure}

Large Language Models (LLMs) have achieved significant progress in recent years, driving remarkable advancements across numerous fields and applications. The popularization of Multimodal Large Language Models (MLLMs)~\cite{li_unimoe,lmeye} has amplified the hallucination~\cite{rawte2023survey} problem, particularly in rich multilingual scenarios.
 This issue is further exacerbated in cross-lingual and cross-modal settings. As shown in Figure~\ref{intro}, even the GPT series models struggle to mitigate hallucinations in these complex scenarios. That is, \textit{MLLMs may yield inconsistent answers when the same factual question is asked to be answered in different languages or presented via different input modalities}.

Factuality benchmarks, such as SimpleQA~\cite{Wei2024MeasuringSF}, have gained increasing attention as effective tools for assessing hallucination in LLMs. These benchmarks use fact-based QA tasks to objectively evaluate model accuracy and reliability. However, most existing benchmarks~\cite{zhang-etal-2025-cchall} focus on textual or visual inputs and are primarily designed for English, lacking coverage of multilingual speech scenarios. As shown in Table~\ref{tab:compare_prior}, a comprehensive benchmark for evaluating multilingual speech settings is still missing.

\begin{table*}[t]
\centering
\small
\setlength{\tabcolsep}{5pt} % Adjust column spacing (default is 6pt)
\begin{tabular}{lccccccccc}

\toprule
{\multirow{2.5}{*}{\textbf{Benchmark} (Text \textcolor{acl_green}{\ding{61}} / {Speech} \textcolor{cvpr_blue}{\ding{71}})}} & \textbf{\multirow{2.5}{*}{Langs.} }& \multirow{2.5}{*}{\shortstack{\textbf{Data} \\ \textbf{Size}}}&\textbf{\multirow{2.5}{*}{Modality} }& \multirow{2.5}{*}{\textbf{Metric}}&\multirow{2.5}{*}{\textbf{Open-ended}}& \multicolumn{4}{c}{\textbf{Support Task}} \\\cmidrule(lr){7-10}

  & && &&& QA & XQA & SQA & XSQA\\ \midrule
   \rowcolor{my_green} TruthfulQA~\cite{lin2022truthfulqa} & English& 817 &\textcolor{acl_green}{\ding{61}}& Acc&{\ding{51}}  &  {\ding{51}} && & \\
 \rowcolor{my_green}  HaluEval~\cite{HaluEval}                          & English        & 5,000  & \textcolor{acl_green}{\ding{61}} & Acc  &      &   {\ding{51}}                    &&&                                          \\
 \rowcolor{my_green}SimpleQA~\cite{Wei2024MeasuringSF}& English&4,326 &\textcolor{acl_green}{\ding{61}}&LLM&{\ding{51}} &{\ding{51}} & & &\\
 \rowcolor{my_green}Chinese SimpleQA~\cite{he2024chinesesimpleqachinesefactuality} & Chinese & 3,000&\textcolor{acl_green}{\ding{61}} &LLM&{\ding{51}}&{\ding{51}} && & \\
\rowcolor{my_green} KoLasSimpleQA~\cite{jiang2025evaluating}& 9&2,147& \textcolor{acl_green}{\ding{61}} &LLM& \ding{51}&{\ding{51}} & && \\ \midrule
\rowcolor{my_blue}SD-QA~\cite{faisal2021sd} & 5& 11,109 & \textcolor{cvpr_blue}{\ding{71}}&F1& && &{\ding{51}}&\\
\rowcolor{my_blue}CompA~\cite{ghosh2024compa} & English& 600 & \textcolor{cvpr_blue}{\ding{71}}&Acc&& &&{\ding{51}}& \\
\rowcolor{my_blue}VoiceBench~\cite{chen2024voicebench} & English& 5,783 & \textcolor{cvpr_blue}{\ding{71}}&Acc / LLM&mix& &&{\ding{51}}& \\
\rowcolor{my_blue}SpeechIQ~\cite{wan2025speechiq} & English& 800 & \textcolor{cvpr_blue}{\ding{71}}&LLM&&& &{\ding{51}}&\\

\rowcolor{my_blue}\textbf{CCFQA} (ours) & 8& 14,400& \textcolor{acl_green}{\ding{61}} / \textcolor{cvpr_blue}{\ding{71}}&F1 / LLM& {\ding{51}} & {\ding{51}} & {\ding{51}}& {\ding{51}}&{\ding{51}}\\ 
\bottomrule
\end{tabular}
\caption{Comparison of CCFQA with Existing Benchmarks. 
CCFQA is a cross-lingual and cross-modal factual benchmark featuring parallel speech-text question pairs across 8 languages, and supporting QA, XQA, SQA, and XSQA tasks.}
\label{tab:compare_prior}
\end{table*}

To bridge this gap and systematically evaluate the factual knowledge consistency of MLLMs in cross-lingual and cross-modal scenarios, we propose a new benchmark named the \textbf{C}ross-lingual and \textbf{C}ross-modal \textbf{F}actuality benchmark (\textbf{CCFQA}), which covers a total of 8 languages. 
The uniqueness of the CCFQA benchmark lies in the fact that each factual question is presented in both textual and spoken input forms, aiming to directly reveal whether the model’s internal factual knowledge exhibits bias under multilingual and multimodal inputs. CCFQA enables a systematic evaluation of MLLMs by measuring the consistency of model responses to the same question across different languages and modalities.
Specifically, we collect a total of 14,400 speech and text QA samples spanning 20 distinct categories. The benchmark supports four task settings: multilingual text QA, cross-lingual text QA (XQA), multilingual spoken QA (SQA), and cross-lingual spoken QA (XSQA). Our systematic evaluation reveals that existing MLLMs exhibit notable inconsistencies in factual knowledge across languages and modalities. Even for simple questions, models often produce contradictory answers when the same query is presented in different languages or modalities, underscoring the difficulty of maintaining factual consistency with diverse inputs.

To address the challenges revealed by this benchmark and improve the factual knowledge consistency of MLLMs, we propose a novel strategy that leverages English as a pivot language to bridge the knowledge gap in cross-lingual question answering. Specifically, we design a simple yet effective end-to-end approach that transforms non-English questions into English, utilizes the strong factual reasoning capabilities of LLMs in English, and then translates the answers back into the target language. We demonstrate that this bridging strategy effectively harnesses the strengths of existing LLMs while reducing the dependency on non-English language resources, significantly enhancing the factual consistency and reliability of MLLMs.

The main contributions are summarized as follows:
\begin{itemize}
\item We release CCFQA, a novel benchmark for evaluating factual question answering in cross-lingual and cross-modal settings, addressing the lack of comprehensive multilingual and multimodal factuality evaluation.

\item We systematically evaluate existing MLLMs and reveal inconsistencies in their answers to the factual questions across different languages and modalities, highlighting the serious challenges in maintaining factual consistency.

\item We design an effective end-to-end strategy that uses English as a bridge language to leverage LLMs’ strong factual knowledge, greatly improving performance and consistency in cross-lingual and cross-modal QA tasks.

\end{itemize}

\section{Related Work}

\subsection{Factuality Benchmarks}

Recently, a series of evaluation benchmarks \cite{liu-etal-2025-projecteval} have emerged in the LLM field, with factuality benchmarks being one of them. Fact-based question answering is a vital research area for evaluating LLMs. TruthfulQA \cite{lin2022truthfulqa}, known as the first fact-based benchmark in the era of LLMs, assesses truthfulness by measuring how LLMs handle questions involving common human misconceptions. Its goal is to prevent models from generating imitative falsehoods learned from their training data. HaluEval \cite{HaluEval} addresses "hallucinations," which are coherent but factually incorrect texts, by using human-annotated samples. SimpleQA \cite{Wei2024MeasuringSF} employs adversarially collected, short, fact-seeking questions with single answers to test a model’s factual recall. Extending fact-based evaluation to other languages, Chinese SimpleQA \cite{he2024chinesesimpleqachinesefactuality} is the first comprehensive Chinese benchmark with high-quality answers, mainly focusing on evaluating Chinese-centered LLMs. Finally, KoLasSimpleQA \cite{jiang2025evaluating} is the first multilingual fact-based benchmark, assessing nine languages on both general and language-specific knowledge.

\subsection{Spoken Question Answering}

Evaluating Spoken Question Answering (SQA) is an evolving field. SD-QA \cite{faisal2021sd} is a typical work addressing dialectal variations with a dataset of over 68,000 spoken prompts, enabling real-world performance and fairness evaluations. CompA \cite{ghosh2024compa}, while not a direct benchmark, offers insights into multisensory fusion relevant for designing multi-modal SQA systems. VoiceBench \cite{chen2024voicebench} provides a comprehensive benchmark for LLM-based voice assistants, assessing general knowledge, instruction-following, and safety under realistic audio conditions. SpeechIQ \cite{wan2025speechiq} introduces a cognition-inspired pipeline, SIQ, which evaluates models across three levels of Bloom's Taxonomy, providing a holistic assessment of a model’s understanding. All of these benchmarks focus on a single voice modality and use non-open-ended answers, highlighting a common limitation in current SQA evaluation methods.

\begin{figure*}[t]
\centering
\includegraphics[width=\linewidth]{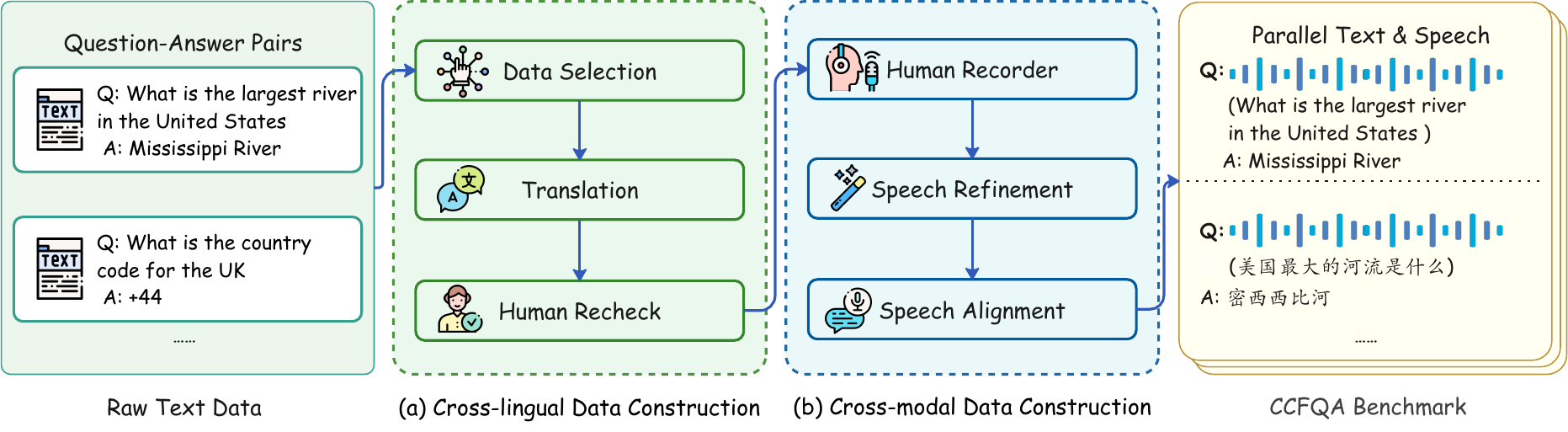}
\caption{CCFQA Dataset Construction: (a) Cross-Lingual Data Construction, (b) Cross-Modal Data Construction.}
\label{dataset}
\end{figure*}

\section{The CCFQA Benchmark}
\subsection{Overview}
This section briefly outlines the construction of the CCFQA benchmark, designed to evaluate MLLMs on multilingual spoken factual question answering. The process consists of two phases: cross-lingual and cross-modal data construction.

\subsection{Cross-lingual Data Construction}
To construct the benchmark, we leverage text-based QA datasets: MKQA~\cite{longpre2021mkqa} and MOOCCubeX \cite{yu2021mooccubex}. 
MKQA provides a wealth of open-domain questions, with the coverage of general knowledge topics such as movies, music, and sports. MOOCCubeX offers questions rooted in educational contexts, covering a wide array of academic subjects.

\paragraph{Data Selection.}
A critical step in our methodology was the manual curation of the initial data pool. We established a rigorous set of exclusion criteria to ensure the utmost quality. Questions were filtered and excluded if they exhibited any of the following characteristics:
\begin{itemize}
    \item \textbf{Ambiguity:} Questions that are poorly phrased, vague, or whose answers may change over time (e.g., "Who is the current prime minister?").
    \item \textbf{Sensitivity:} Content containing personally identifiable information (PII), offensive language, or highly controversial political or social topics.
    \item \textbf{Factual Incorrectness:} Pairs where the provided answer was outdated, disputed, or verifiably incorrect.
    \item \textbf{Culture Independent:} Questions whose answers are not universally factual but vary across different cultural or legal contexts. (e.g., "What is the age of consent?")
\end{itemize}

\paragraph{Translation.}
Subsequently, the filtered questions and their corresponding canonical answers were professionally translated into other seven target languages. To achieve high-quality translation, we utilized the advanced machine translation capabilities of GPT-4.1. Each translation prompt was carefully engineered to preserve the factual nature of the query and the accuracy of the answer, explicitly instructing the model to maintain semantic equivalence and formal tone. 
\begin{table*}[t]
\centering
\includegraphics[width=\linewidth]{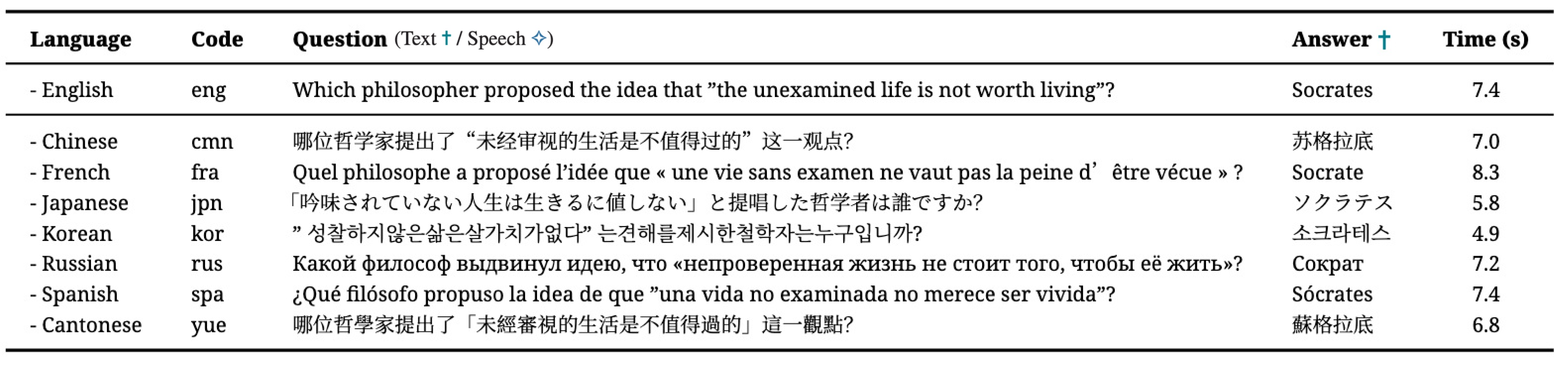}
\caption{Questions and Answers in All Supported Languages for One Instance in CCFQA.}
\label{tbl:answer-examples}
\end{table*}

\paragraph{Human Recheck}  
We performed manual verification for certain languages (Chinese, English, Japanese) and employed back-translation followed by review for the remaining languages to ensure the accuracy and consistency of all questions and answers.

\subsection{Cross-modal Data Construction}
The final phase involves converting the verified text-based QA pairs into a high-quality speech dataset for evaluating MLLMs' speech processing. This includes guided human recordings, audio enhancement, and ASR-based quality checks to identify and re-record low-quality samples. This iterative process ensures the speech data accurately matches the text.

\paragraph{Human Recording.}  
We recruit a balanced and diverse group of native speakers (including male and female) for the 8 target languages to record the \textbf{parallel speech-text samples}. Each volunteer is instructed to read the sentences with clear, natural enunciation, at a consistent pace, and in a neutral tone, avoiding emotional inflections or disfluencies. This protocol is designed to produce high-quality, clean audio recordings ideal for MLLM evaluation.

\paragraph{Speech Refinement.}  
To enforce a final layer of quality control on the audio data, we implement a systematic speech refinement loop. We apply audio augmentation to enhance samples with low volume levels. Additionally, we employ Whisper-large-v3~\cite{whisper}, an Automatic Speech Recognition (ASR) model, for a comprehensive quality check. The ASR system transcribes every recorded audio file. We then calculate the Word Error Rate (WER) and Character Error Rate (CER) by comparing the ASR-generated transcripts with the original ground-truth text.

\paragraph{Speech Alignment.}
Any audio file exhibiting a WER above a predefined low threshold is automatically flagged as having a potential mispronunciation or recording artifact. Our volunteers are then asked to re-record these specific flagged utterances. This iterative process of recording, ASR-based verification, and re-recording ensures that the final audio dataset is of exceptional quality, accurately reflects the underlying text, and provides a reliable basis for cross-modal evaluation for MLLMs.

\newpage

\subsection{Benchmark Statistics}
This section introduces the CCFQA benchmark, a novel and comprehensive resource designed for evaluating the knowledge consistency of MLLMs across various QA tasks. Specifically, CCFQA supports the evaluation of QA, XQA, SQA, and XSQA for MLLMs, making it a versatile tool for multimodal and multilingual research.

\begin{figure}[t]
\centering
\includegraphics[width=0.95\linewidth]{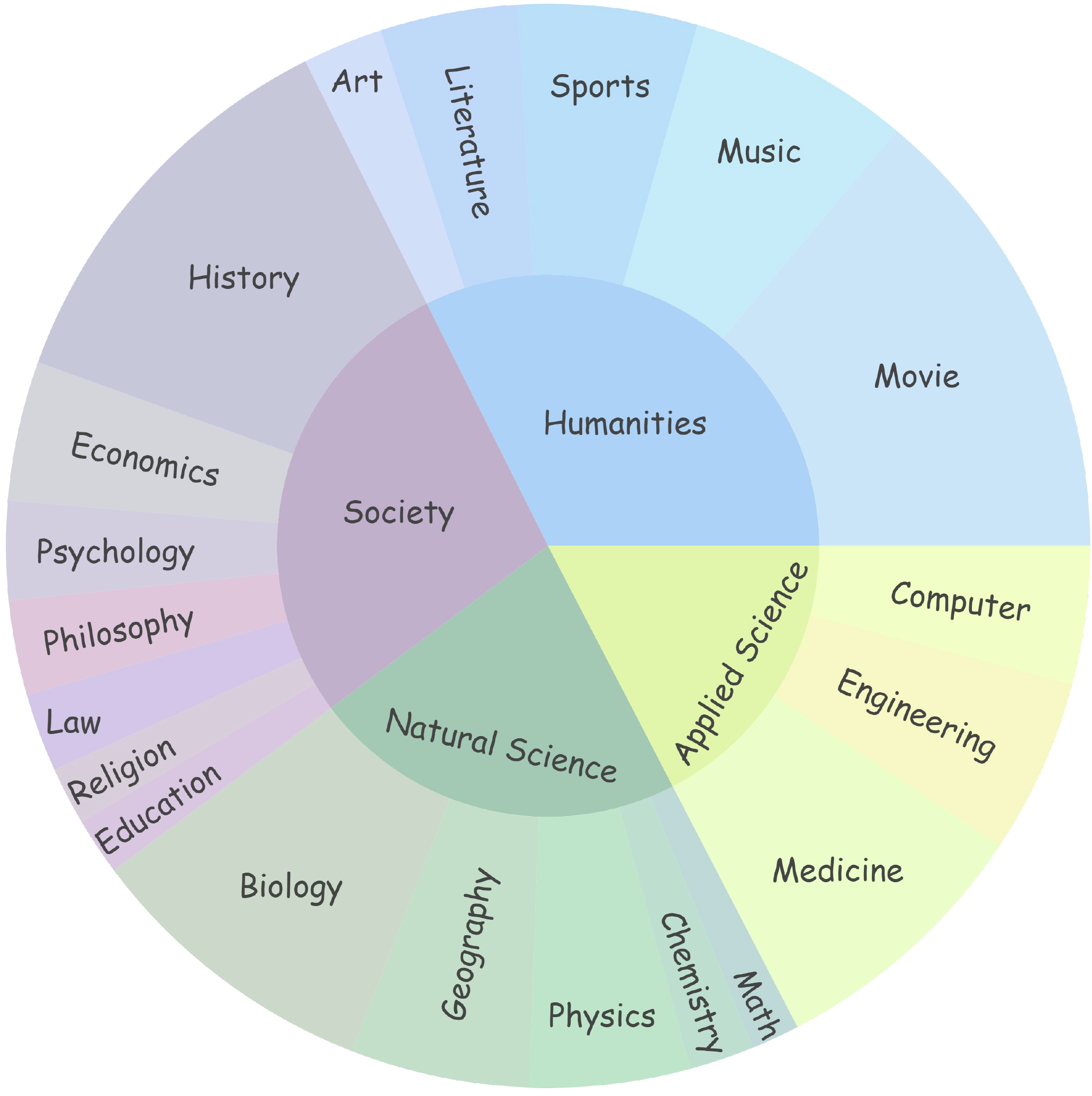}
\caption{The Question Categories in the CCFQA.}
\label{domain}
\end{figure}

\begin{table}[t]
\centering
\includegraphics[width=\linewidth]{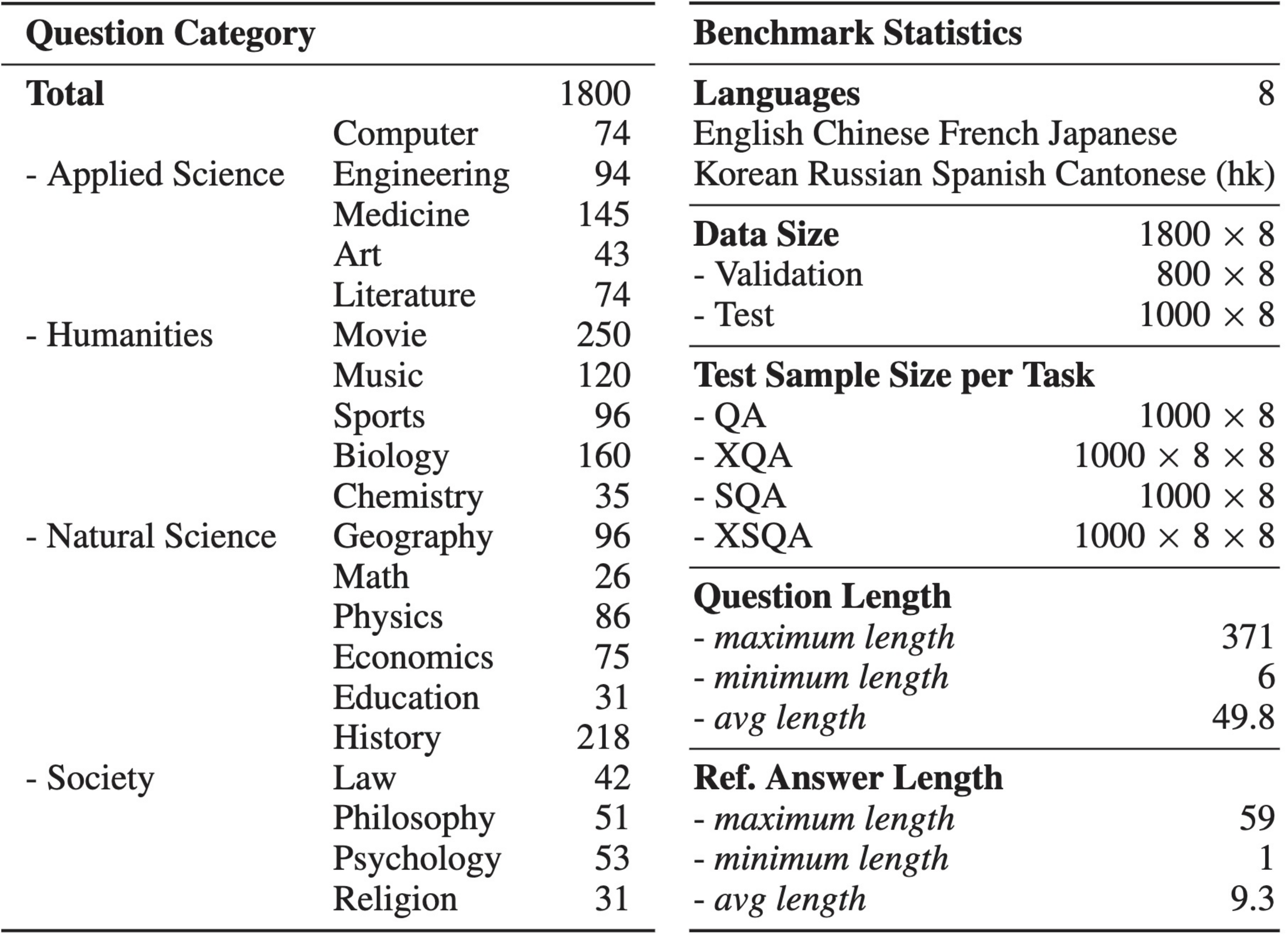}
\caption{Benchmark Statistics. CCFQA supports the evaluation of QA, XQA, SQA, and XSQA for MLLMs. It features parallel speech-text question pairs across 8 languages. Due to the cross-lingual evaluation settings in XQA and XSQA, a full evaluation of one MLLM requires 128,000 individual requests.}
\label{tbl:answer-examples}
\end{table}

\paragraph{Language \& Domain Distribution.}
As illustrated in Figure \ref{domain}, our dataset is meticulously constructed from content spanning a total of 8 languages with 4 major domains (humanities, society, natural science and applied science), which are further categorized into 20 sub-domains. 
\paragraph{Test \& Validation Set Composition.}
The CCFQA benchmark contains a total of \textbf{1800} parallel speech-text question pairs covering 8 languages (each question available in both text and speech forms). 
Among them, \textbf{800} pairs are used for validation and \textbf{1000} pairs for testing. 
Due to the cross-lingual evaluation settings in XQA and XSQA, a full evaluation of one MLLM requires \textbf{128,000} individual requests.

\subsection{Cross-Lingual and Cross-Modal Consistency}
The CCFQA benchmark is designed to evaluate the consistency of MLLMs across different languages and input modalities. 
A robust MLLM should give the same answer to an identical factual question, regardless of the input language or whether the question is presented as text or speech. 
However, most existing benchmarks lack the fully parallel data needed for such an evaluation, especially in multilingual spoken scenarios. 
To address this gap, CCFQA provides a fully parallel speech-text dataset covering eight languages. 
This enables systematic assessment of:
\begin{itemize}    \item \textbf{Cross-lingual Consistency:} Can the model produce equivalent answers across multiple languages?
\item \textbf{Cross-modal Consistency:} Can the model maintain answer quality across text and speech inputs?\end{itemize}
Thanks to this parallel design, evaluations can be controlled and directly compared, revealing modality and language biases that are often hidden in non-parallel benchmarks.

\begin{figure}[t]
\centering
\includegraphics[width=\linewidth]{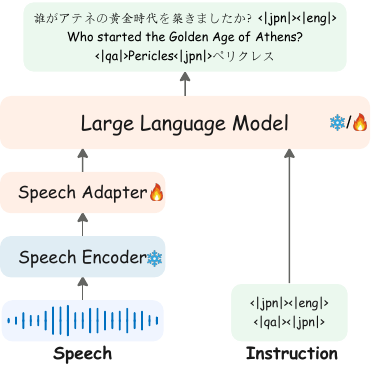}
\caption{The Architecture of LLM-SQA. }
\label{framework}
\end{figure}

\label{sec:mllm}

\section{Few-Shot Transfer Learning for LLM-SQA}

\paragraph{Pretraining.}  
Following the design of LLM-SRT~\cite{du2025makingllmsbettermanytomany}, we adopt a sequential curriculum learning strategy that trains the model on three tasks in order: (1) Automatic Speech Recognition (ASR), (2) Speech Recognition and Translation (SRT), and (3) Spoken Question Answering (SQA). To help the model distinguish between different tasks, we design explicit task instructions that ensure clear separation between instructions and predicted outputs, as shown in Figure \ref{framework}. The ASR and SRT tasks are trained on the FLEURS dataset~\cite{conneau2023fleurs}.

\paragraph{SQA Few-shot Learning.}  
We adopt a two-stage training strategy for the SQA task. In the first stage, we perform supervised fine-tuning using approximately 3,000 synthetic English speech samples paired with corresponding text. This enables the model to learn the task structure and factual question-answering ability in a high-resource setting. 

In the second stage, we enable cross-lingual transfer by applying a \textbf{5-shot} few-shot learning setup on target languages. With only a handful of annotated examples, the model quickly adapts to new languages and modalities. By leveraging English as a bridge, the learned knowledge is effectively transferred to other languages (e.g., Japanese or Cantonese), greatly reducing the reliance on large-scale annotated data in each target language. This strategy enables our model to perform cross-lingual spoken question answering across \textbf{eight languages}, despite minimal supervision.

\paragraph{Language and Task Support.}
Our MLLM is pre-trained from scratch based on GemmaX2-9B~\cite{cui2025multilingual}, with added support for Cantonese (Hong Kong SAR). Our model supports ASR, Speech-to-Text Translation (S2TT), and SQA tasks together within a single query.

\begingroup
\renewcommand{\arraystretch}{1.25 } % 将行间距增加到原来的 1.5 倍
\begin{table}[t]
  \centering
  \small
  \resizebox{1.0\linewidth}{!}{
  \setlength{\tabcolsep}{1pt} % Adjust column spacing (default is 6pt)
    \begin{tabular}{c c c c} 
    \toprule 

\textbf{Task}&\textbf{Instruction}&\textbf{Prediction} \\ 
\hline
 \multirow{3}{*}{ASR} & \texttt{<|eng|>}  &  what is the largest city in the uk\\ 

 & \texttt{<|fra|>}  &  quelle est la plus grande ville du royaume-uni\\ 
 & \texttt{<|spa|>}  & cual es la ciudad mas grande de uk \\
\hline
\multirow{4}{*}{SRT} &\multirow{2}{*}{\texttt{\small<|eng|><|fra|>}} & what is the largest city in the uk\texttt{\small<|eng|><|fra|>}\\ 
&&quelle est la plus grande ville du royaume-uni\\
 &\multirow{2}{*}{\texttt{\small<|spa|><|eng|>}} & cual es la ciudad mas grande de uk\texttt{<|spa|><|eng|>}\\ 
&&what is the largest city in the uk\\ \hline
\multirow{9}{*}{SQA}& \multirow{2}{*}{\texttt{\small<|qa|>}}& what is the largest city in the uk
\\
& & \texttt{<|qa|>}London  \\ \cmidrule(r){3-3}
 &  \multirow{2}{*}{\texttt{\small<|spa|><|eng|>}} & cual es la ciudad mas grande de uk\texttt{<|spa|><|eng|>}\\
& \multirow{2}{*}{\texttt{\small<|qa|>}}& what is the largest city in the uk
\\
& & \texttt{<|qa|>}London \\ \cmidrule(r){3-3}
& \multirow{2}{*}{\texttt{\small<|spa|><|eng|>}}  & cual es la ciudad mas grande de uk\texttt{<|spa|><|eng|>}\\ 

&\multirow{2}{*}{\texttt{\small<|qa|><|fra|>} } & what is the largest city in the uk
\\
&& \texttt{<|qa|>}London\texttt{<|fra|>}Londres  \\

\bottomrule
    \end{tabular}}
\caption{Instruction Design. LLM-SQA consists of a speech encoder, speech adapter, and LLM. A curriculum learning strategy sequentially trains the ASR, SRT, and SQA tasks. We train the \texttt{<|qa|>} task in English, and for other languages, cross-lingual knowledge transfer can be achieved with only 5-shot trainings, using English as a bridge.}
\label{pattern}
\end{table}%
\endgroup

\section{Experiments}
\subsection{Experiment Setting}

\paragraph{Model Architecture.}  
Our MLLM includes a frozen speech encoder, a trainable adapter composed of a Q-Former with 80 queries of dimension 768 and an MLP, and an LLM, as detailed in Table~\ref{tab:parameters}.

\paragraph{Training Details.}  
Experiments run for one week on four A100 (80GB) GPUs. We use AdamW with a peak learning rate of $1 \times 10^{-4}$, 1000-step warm-up, and linear decay thereafter.

\paragraph{Comparing MLLMs.}  
We compare five MLLMs: GPT-4o-mini~\cite{achiam2023gpt}, Phi-4-Multimodal~\cite{abouelenin2025phi}, Qwen2-Audio~\cite{chu2024qwen2}, and the Qwen2.5-Omni series~\cite{xu2025qwen2}.

\paragraph{Evaluation Metrics.}  
We evaluate using the F1 score and an LLM judge. Details on the judge selection and experimental setup are in the Appendix.

\begin{table}[h] 
\centering 
\footnotesize 
\setlength{\tabcolsep}{1.4mm} 
\begin{tabular}{lrcl} 
\toprule
\multirow{2}{*}{\textbf{Modules}} & \multirow{2}{*}{\textbf{Param}} & \textbf{Training} & \multirow{2}{*}{\textbf{Details}} \\
& & \textbf{stage} & \\
\midrule
Speech Encoder & $\sim$635M & - & Whisper's encoder \\
\midrule
Speech Adapter & {\color{cvpr_blue}$\sim$80.5M} & all & Q-Former and MLP \\
\midrule
LLM & $\sim$9.2B & - & GemmaX2-9B \\
\midrule
LLM adapter & {\color{cvpr_blue}$\sim$8.9M} & II\&III & LoRA (r=16, alpha=32) \\
\midrule
Total & $\sim$10B & & \\ 
\bottomrule
\end{tabular}
\caption{MLLM Training Settings. The blue color indicates the number of trainable parameters. }
\label{tab:parameters} 
\end{table}

\begin{table*}[h]
\centering
\small
\renewcommand{\arraystretch}{0.92} % 设置行间距为默认的1.2倍
\setlength{\tabcolsep}{4pt} % Adjust column spacing (default is 6pt)
\begin{tabular}{lccccccccc} \toprule 
\multirow{2}{*}{F1 / LLM Acc $\uparrow$} &\multicolumn{9}{c}{\textbf{CCFQA}  (Text \textcolor{acl_green}{\ding{61}} / {Speech} \textcolor{cvpr_blue}{\ding{71}})}\\ 

& Cmn & Eng & Fra & Jpn & Kor & Rus& Spa & Yue & Avg. \\ \midrule 
\multicolumn{10}{c}{\textcolor{acl_green}{\ding{61}} QA (X$\rightarrow$X)}  \\ \midrule
\rowcolor{my_green}GPT-4o-mini  & \textbf{59.2} / \textbf{63.6} & \textbf{78.7} / \textbf{82.0} & \textbf{74.1} / \textbf{73.7} & \textbf{60.3} / \textbf{63.4} & \textbf{50.9} / \textbf{55.3} & \textbf{62.7} / \textbf{42.0} & \textbf{74.1} / \textbf{76.3} & \textbf{51.6} / \textbf{59.0} & \textbf{63.9} / \textbf{64.4}\\
\rowcolor{my_green}Phi-4-Multimodal & 6.9 / 13.4 & 32.4 / 28.7 & 32.6 / 15.8 & 6.5 / 11.9 & 7.1 / 5.6 & 22.4 / 11.3 & 32.0 / 16.6 & 4.0 / 6.7 & 18.0 / 13.8\\
\rowcolor{my_green}Qwen2-Audio & 17.7 / 40.5 & 46.5 / 54.4 & 46.0 / 37.2 & 9.9 / 18.6 & 10.3 / 11.4 & 37.3 / 18.2 & 45.8 / 39.9 & 9.7 / 24.4 & 27.9 / 30.6\\
\rowcolor{my_green}Qwen2.5-Omni-3B  & 13.4 / 18.1 & 29.7 / 26.7 & 33.0 / 13.9 & 7.0 / 2.0 & 4.7 / 2.5 & 27.1 / 4.3 & 37.0 / 12.8 & 13.2 / 13.8 & 20.6 / 11.8\\
\rowcolor{my_green}Qwen2.5-Omni-7B  & 45.3 / 53.2 & 66.4 / 60.3 & 58.1 / 39.7 & 38.1 / 32.9 & 25.1 / 19.9 & 48.1 / 22.7 & 58.9 / 43.7 & 31.7 / 33.2 & 46.5 / 38.2\\

 \midrule
 \multicolumn{10}{c}{\textcolor{acl_green}{\ding{61}} XQA (X$\rightarrow$8)}  \\ \midrule
\rowcolor{my_green}GPT-4o-mini  & \textbf{60.3} / \textbf{63.1} & \textbf{64.4} / \textbf{68.8} & \textbf{62.1} / \textbf{65.0} & \textbf{57.2} / \textbf{59.4} & \textbf{56.1} / \textbf{56.6} & \textbf{59.2} / \textbf{58.7} & \textbf{61.7} / \textbf{65.8} & \textbf{56.6} / \textbf{60.0} & \textbf{59.7} / \textbf{62.2}\\
\rowcolor{my_green}Phi-4-Multimodal & 17.8 / 15.5 & 19.1 / 19.9 & 18.5 / 16.0 & 17.8 / 15.2 & 17.7 / 10.0 & 17.6 / 15.7 & 18.6 / 16.4 & 16.8 / 13.3 & 18.0 / 15.3\\
\rowcolor{my_green}Qwen2-Audio & 25.5 / 22.3 & 25.1 / 21.9 & 24.1 / 18.6 & 23.2 / 18.1 & 23.1 / 13.9 & 24.2 / 17.4 & 24.3 / 19.9 & 23.9 / 21.1 & 24.2 / 19.2\\
\rowcolor{my_green}Qwen2.5-Omni-3B  & 12.5 / 11.1 & 18.6 / 16.9 & 17.9 / 10.7 & 8.4 / 2.8 & 10.5 / 3.6 & 13.6 / 6.2 & 18.4 / 10.6 & 11.6 / 9.4 & 13.9 / 8.9\\
\rowcolor{my_green}Qwen2.5-Omni-7B  & 45.7 / 42.4 & 47.9 / 39.7 & 43.0 / 35.8 & 41.7 / 32.7 & 37.4 / 26.9 & 40.7 / 31.7 & 43.5 / 35.4 & 36.5 / 31.4 & 42.1 / 34.5\\

 \midrule
 
 \multicolumn{10}{c}{\textcolor{cvpr_blue}{\ding{71}} SQA (X$\rightarrow$X)}  \\ \midrule
\rowcolor{my_blue}GPT-4o-mini-Audio  & 38.6 / 36.6 & \textbf{75.9} / \textbf{74.7} & 54.0 / \textbf{40.4} & 42.9 / \textbf{39.5} & 31.8 / 28.9 & 53.9 / 28.2 & 63.7 / \textbf{57.1} & 21.1 / 17.8 & 47.7 / \textbf{40.4}\\
\rowcolor{my_blue}Phi-4-Multimodal & 7.1 / 25.3 & 40.0 / 56.5 & 39.1 / 34.4 & 9.1 / 18.9 & 1.7 / 1.2 & 10.5 / 0.4 & 38.3 / 37.7 & 2.4 / 1.6 & 18.5 / 22.0\\
\rowcolor{my_blue}Qwen2-Audio & 19.4 / 31.5 & 48.6 / 31.3 & 44.7 / 19.2 & 8.1 / 5.0 & 8.9 / 2.6 & 38.0 / 5.6 & 46.1 / 24.6 & 8.0 / 16.3 & 27.7 / 17.0\\
\rowcolor{my_blue}Qwen2.5-Omni-3B  & 41.3 / 39.6 & 59.1 / 41.4 & 35.7 / 15.3 & 21.1 / 11.4 & 17.2 / 7.9 & 36.9 / 9.3 & 48.1 / 23.2 & 19.4 / 19.1 & 34.9 / 20.9\\
\rowcolor{my_blue}Qwen2.5-Omni-7B  & \textbf{55.2} / \textbf{53.5}	&68.2 / 58.5&	43.2 / 25.7&	33.0 / 25.7	&21.4 / 11.5	&46.0 / 21.6	&54.6 / 37.2	&\textbf{30.7} / \textbf{31.7}	&44.0 / 33.2
\\
\rowcolor{my_blue}LLM-SQA (ours) & 51.3 / 45.5 & 74.8 / 60.5 & \textbf{60.3} / 36.8 & \textbf{43.4} / 38.8 & \textbf{41.6} / \textbf{36.1} & \textbf{54.1} / \textbf{32.5} & \textbf{64.5} / 45.3 & 25.9 / 27.0 & \textbf{52.0} / 40.3\\
 \midrule
  \multicolumn{10}{c}{ \textcolor{cvpr_blue}{\ding{71}} XSQA (X$\rightarrow$8) } \\ \midrule
\rowcolor{my_blue}GPT-4o-mini-Audio  & 44.9 / 34.8 & \textbf{62.9} / \textbf{63.7} & 42.7 / 36.8 & 46.1 / 37.3 & 40.7 / 29.9 & 48.0 / \textbf{39.9} & 53.0 / \textbf{50.0} & 26.9 / 16.5 & 45.7 / 38.6\\
\rowcolor{my_blue}Phi-4-Multimodal & 20.7 / 6.7 & 23.3 / 10.4 & 22.6 / 10.7 & 20.6 / 5.6 & 21.5 / 0.6 & 18.4 / 0.2 & 22.3 / 10.6 & 18.7 / 0.5 & 21.0 / 5.7\\
\rowcolor{my_blue}Qwen2-Audio & 25.9 / 16.6 & 26.5 / 17.2 & 23.5 / 10.2 & 23.7 / 8.1 & 21.2 / 2.3 & 23.2 / 5.8 & 24.4 / 12.3 & 24.4 / 13.6 & 24.1 / 10.7\\
\rowcolor{my_blue}Qwen2.5-Omni-3B  & 34.6 / 24.1 & 42.6 / 27.3 & 26.1 / 13.3 & 26.6 / 13.3 & 25.8 / 12.9 & 26.7 / 12.9 & 35.1 / 19.3 & 20.6 / 13.9 & 29.8 / 17.1\\
\rowcolor{my_blue}Qwen2.5-Omni-7B  & 47.6 / 38.8&
48.1 / 45.3&
32.8 / 23.5&
39.3 / 26.2&
34.5 / 21.0&
37.7 / 24.9&
42.1 / 33.6&
26.1 / 22.6&
38.5 / 29.5
\\
\rowcolor{my_blue}LLM-SQA (ours) & \textbf{56.6} / \textbf{45.8} & 48.9 / 40.3 & \textbf{49.9} / \textbf{38.6} & \textbf{52.2} / \textbf{39.9} & \textbf{52.2} / \textbf{40.9} & \textbf{52.3} / 39.6 & \textbf{54.4} / 44.9 & \textbf{44.9} / \textbf{27.7} & \textbf{51.4} / \textbf{39.7}\\

\bottomrule

\end{tabular}

\caption{
F1 and LLM-based accuracy of MLLMs on the CCFQA benchmark, including cross-lingual (QA $\rightarrow$ XQA / SQA $\rightarrow$ XSQA) and cross-modal (QA $\rightarrow$ SQA / XQA $\rightarrow$ XSQA) settings. 
Performance degradation is observed in both cross-lingual and cross-modal scenarios. 
% Full results are provided in Tables~\ref{tab:text} and~\ref{tab:speech} in the Appendix.
}
\label{tab:main}
\end{table*}

\subsection{Main Results}

As shown in Table~\ref{tab:main}, we evaluate MLLMs across linguistic (monolingual vs. cross-lingual) and modalities (text vs. speech) settings, with a focus on consistency in factual knowledge. We have the following insightful observations:

\paragraph{Text-based QA and XQA.}
In text-based tasks, GPT-4o-mini achieves the best performance. Among open-source MLLMs, Qwen2.5-Omni-7B obtains the highest score. In terms of language-wise performance, English (eng), French (fra), and Spanish (spa) achieve relatively high scores, while Korean (kor) and Cantonese (yue) perform relatively worse.

\paragraph{Speech-based SQA and XSQA.}
In speech-based tasks, GPT-4o-mini-Audio achieves the best performance in English (eng), while Qwen2.5-Omni-7B performs best in Mandarin (cmn) and Cantonese (yue). LLM-SQA demonstrates strong overall performance through a knowledge transfer strategy based on few-shot learning.

\paragraph{Comparison Between F1 / LLM Scores.}
We observe differences between F1 and LLM-based accuracy scores. Low F1 with high accuracy suggests the MLLM knows the facts but struggles to follow instructions (prompt requires answers without explanations). Conversely, high F1 but low accuracy indicates fluent, content-rich answers that are related to the question but factually incorrect, reflecting hallucinations.

\begin{table}[t] 
\centering 
\footnotesize 
\setlength{\tabcolsep}{1.4mm} 
\begin{tabular}{lcc} 
\toprule
\multirow{1}{*}{Consistency $\uparrow$} & \multirow{1}{*}{\textbf{Cross-Lingual}} & \multirow{1}{*}{\textbf{Cross-Modal}}  \\\midrule
GPT-4o-mini & \textbf{96.6} / \textbf{95.5}&62.7 / 62.1\\
Phi-4-Multimodal& 90.8 / 25.9& 62.7 / 37.5\\
Qwen2-Audio &62.4 / 62.9& 55.6 / 56.0\\
Qwen2.5-Omni-3B& 75.4 / 81.8& 56.5 / 52.0\\
Qwen2.5-Omni-7B &90.3 / 87.2& \textbf{90.3} / \textbf{85.5}\\

\bottomrule
\end{tabular}
\caption{Cross-Lingual and Cross-Modal Consistency. Scores are the ratio of the performance across the four tasks.}
\label{tab: consistency} 
\end{table}

\paragraph{Cross-Lingual Challenge.} As shown in Table \ref{tab: consistency}, in cross-lingual settings (XQA and XSQA), most models (except for the GPT-4o-mini) exhibit a significant performance drop compared to their results on QA and SQA tasks. This highlights the difficulty of multilingual alignment, as models generally achieve their best performance in English but perform poorly on low-resource languages.

\paragraph{Cross-Modal Challenge.} As shown in Table \ref{tab: consistency}, a pronounced performance degradation is observed across most models when the input modality shifts from text to speech. This "modality gap" highlights the inherent challenges in multimodal alignment. Notably, Qwen2.5-Omni-7B exhibits superior cross-modal consistency, highlighting the effectiveness of the Omni design.

\subsection{Further Analysis}
\begin{figure}[t]
\centering
\includegraphics[width=\linewidth]{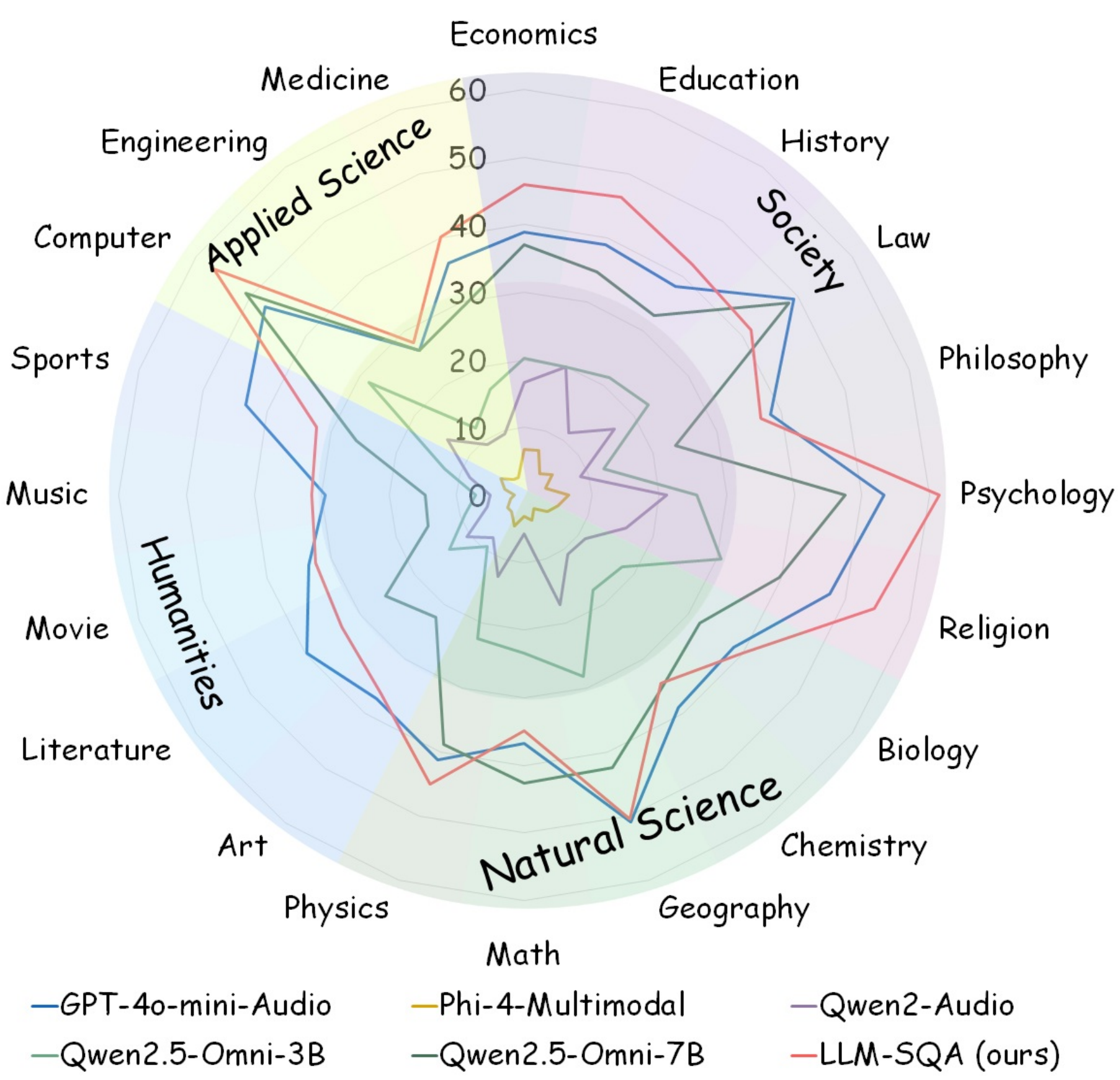}
\caption{Performance Across Categories on XSQA Task.}
\label{output}
\end{figure}
\paragraph{Category Performance on XSQA.}
Figure~\ref{output} presents a radar chart illustrating model performance across 20 knowledge categories on the XSQA task. We observe that LLM-SQA demonstrates the most balanced and consistently high performance across both scientific (e.g., Biology, Chemistry, Medicine) and social domains (e.g., Psychology, History, Education), significantly outperforming baselines. In contrast, other models such as Qwen2.5-Omni-7B show more varied performance, excelling in specific areas like Computer and Law but underperforming in fields such as Movie, Music and Philosophy. 

\paragraph{Speech Length Analysis.}
We evaluate SQA performance across different speech durations to assess robustness to input length. LLM-SQA achieves the highest average accuracy (40.3), surpassing all open-source baselines and matching GPT-4o-mini-Audio (40.4). It performs especially well on short (0–5s) and medium (5–10s) segments, where models like Qwen2-Audio and Phi-4-Multimodal show notable drops. This demonstrates LLM-SQA’s strength in handling brief spoken queries common in real-world use. In contrast, while Qwen2.5-Omni-7B performs decently overall (34.5), it lags behind LLM-SQA across all length ranges, highlighting the effectiveness of our instruction-tuning strategy.

\paragraph{Language Tags for MLLMs.}  
Most current MLLMs are built upon base LLMs and extend their capabilities to support multimodal inputs by incorporating special tokens. The design of these special tokens plays a crucial role in effectively distinguishing different languages and tasks. For example, phi-4-multimodal employs a special token sequence for SQA such as \texttt{<|user|><|audio\_1|><|end|><|assistant|>}. While this design works well for standard SQA, it struggles with XSQA, causing a sharp drop in performance. This shows that the design of special tokens needs to carefully consider language support.

\paragraph{Speech Quality Check.}  
The table \ref{tab:wer} shows the Word Error Rate (WER) and Character Error Rate (CER) in our benchmark. English (Eng) and Mandarin (Cmn) perform best, with WER and CER of 3.2 and 6.8. Some languages like French (Fra), Russian (Rus), and Cantonese (Yue) have higher error rates mainly due to many domain-specific proper nouns that are harder for ASR model to recognize, even when pronounced correctly. Overall, these results indicate that the speech data in our benchmark is of high quality with generally low error rates across languages.
% \subsection{Comparative Model Analysis}

\begin{table}[t] 
\centering 
\footnotesize 
\setlength{\tabcolsep}{1.4mm} 
\begin{tabular}{lcccc} 
\toprule
\multirow{1}{*}{LLM Acc $\uparrow$} & \multirow{1}{*}{\textbf{0-5s}} & \multirow{1}{*}{\textbf{5-10s}} & \multirow{1}{*}{\textbf{10-30s}} & \multirow{1}{*}{\textbf{Avg.}} \\\midrule
  \multicolumn{5}{c}{ \textcolor{cvpr_blue}{\ding{71}} SQA (X$\rightarrow$X) } \\ \midrule
\rowcolor{my_blue}GPT-4o-mini-Audio & 38.6 &\textbf{42.6}&\textbf{40.1}&\textbf{40.4}\\
\rowcolor{my_blue}Phi-4-Multimodal &17.7 & 26.4 & 27.4&22.0\\
\rowcolor{my_blue}Qwen2-Audio & 14.7 & 19.5&18.6&17.0\\
\rowcolor{my_blue}Qwen2.5-Omni-3B &20.5&21.7&18.6&20.9\\
\rowcolor{my_blue}Qwen2.5-Omni-7B &32.1 &34.9 & 29.0&33.2\\
\rowcolor{my_blue}LLM-SQA (ours)& \textbf{40.3}&40.8&36.1&40.3 \\

\bottomrule
\end{tabular}
\caption{LLM-based Accuracy Across Speech Lengths.}
\label{tab:length} 
\end{table}

\begin{table}[t] 
\centering 
\footnotesize 
\setlength{\tabcolsep}{1.4mm} 
\begin{tabular}{cccc|cccc} 
\toprule
\multicolumn{4}{c}{WER $\downarrow$} &\multicolumn{4}{c}{CER $\downarrow$}\\
\multirow{1}{*}{Eng} & \multirow{1}{*}{Fra} & \multirow{1}{*}{Rus} & \multirow{1}{*}{Spa} &Cmn& Jpn&Kor&Yue\\\midrule
  3.2&13.8&18.2&9.9&6.8&8.7&8.2&16.8\\

\bottomrule
\end{tabular}
\caption{Speech Quality Check. We use Whisper-Large-v3 to evaluate ASR error rates.} 
\label{tab:wer} 
\end{table}

\section{Conclusion}
This paper proposes a factual benchmark, CCFQA, for evaluating the consistency of MLLMs in cross-lingual and cross-modal scenarios. The benchmark covers 8 languages and consists of parallel, real human speech and text data. In addition, we introduce a simple yet effective few-shot learning strategy that leverages English as a bridge language for cross-lingual knowledge transfer.
\section{Limitations}
Although this paper explores cross-lingual and cross-modal consistency between speech and text inputs, the current benchmark remains limited to these two modalities. Future work could consider extending it to additional modalities (e.g., vision). Furthermore, the few-shot method proposed in this paper, while enhancing multilingual capabilities, also introduces a language bias centered around English.
\section{Ethical Statement} % Or \section{Risk and Ethical Assessment}
We emphasize the importance of ethics in research involving speech data. All volunteers have signed a Voice Authorization License Agreement, granting permission for their recorded speech to be used for research purposes. The data usage strictly complies with applicable privacy and data protection regulations. We are committed to handling all data in a responsible manner and to safeguarding the confidentiality of participants throughout the entire project.
\section*{Acknowledgements}
The research in this article is supported by the National Science and Technology Major Program (Grant No. 2024ZD01NL00101), the National Science Foundation of China (U22B2059, 62276083, 62506182), the 5G Application Innovation Joint Research Institute’s Project (A003), and the Major Key Project of PCL (Grant No. PCL2025A12, PCL2025A03).

\bibliography{aaai2026}

\clearpage
\appendix
\section{Appendix}
\subsection{Language Support}

Our benchmark supports a diverse set of languages spanning multiple language families and scripts, as summarized in Table~\ref{tab:all_languages}. Specifically, we cover eight languages, including high-resource languages such as Mandarin Chinese (cmn), English (eng), French (fra), Japanese (jpn), Korean (kor), Russian (rus), and Spanish (spa). These languages belong predominantly to the Sino-Tibetan, Indo-European, Japonic, and Koreanic families.

Additionally, we include Yue Chinese (yue), a low-resource variety primarily spoken in the Hong Kong SAR, written in the traditional Chinese script. This inclusion demonstrates the model's capability to handle both widely spoken and regional dialects with varying resource availability. The comprehensive language support ensures that our benchmark is applicable across a broad linguistic spectrum, facilitating robust and fair assessments of multilingual language models.

\begin{table}[h]  
\small
\centering
\setlength{\tabcolsep}{4pt} % Adjust column spacing (default is 6pt)

\begin{tabular}{ccccccccc}
\toprule
\textbf{Code} & \textbf{Language} & \textbf{Family}  & \textbf{Script}& \textbf{Resource}  \\
\midrule
cmn & Mandarin Chinese & Sino-Tibetan  & Hans &High \\
eng & English & Indo-European  & Latn &High \\
fra & French & Indo-European  & Latn &High\\
jpn & Japanese & Japonic  & Jpan &High \\
kor & Korean & Koreanic  & Kore &High \\
rus & Russian & Indo-European  & Cyrl &High  \\
spa & Spanish & Indo-European & Latn &High \\
yue & Yue Chinese & Sino-Tibetan  & Hant  &Low\\
\bottomrule
\end{tabular}

\caption{Language Support. Language codes follow ISO 639-3. Yue Chinese here is the variety spoken in the Hong Kong SAR.
}
\label{tab:all_languages}
\end{table}

\subsection{Evaluation Model}
Specifically, Gemma3-27B \cite{team2025gemma} is selected as the judge LLM, which, given a question, an answer, the language direction, and a response, judges whether the response is correct by answering "yes" or "no." For the same task, all models are evaluated using identical prompts and consistent sampling parameters set via vLLM \cite{kwon2023efficient}. We selected the LLM as our evaluation model based on the following key factors:

\begin{itemize}
    \item \textbf{Performance:} A fundamental principle is that the evaluation model should outperform the baselines. As shown in Table~\ref{tab:gemma3}, Gemma3-27B achieves better results than all compared baselines, ensuring accurate and trustworthy evaluation outcomes.

    \item \textbf{Comparability:} Gemma3-27B can be deployed locally, providing a consistent evaluation environment and reproducible results. This eliminates the uncertainties introduced by frequent API updates and ensures fair comparisons across different models.

    \item \textbf{Efficiency:} Our experimental setup requires evaluating each MLLM on up to 128,000 entries. Using API-based models such as GPT-4o would lead to unacceptable latency and strict daily request limits. By integrating Gemma3-27B with vLLM, we enable fast inference, significantly improving efficiency and reducing both the time and cost of large-scale evaluation.
\end{itemize}

\begin{figure*}[h]
\centering
\includegraphics[width=\linewidth]{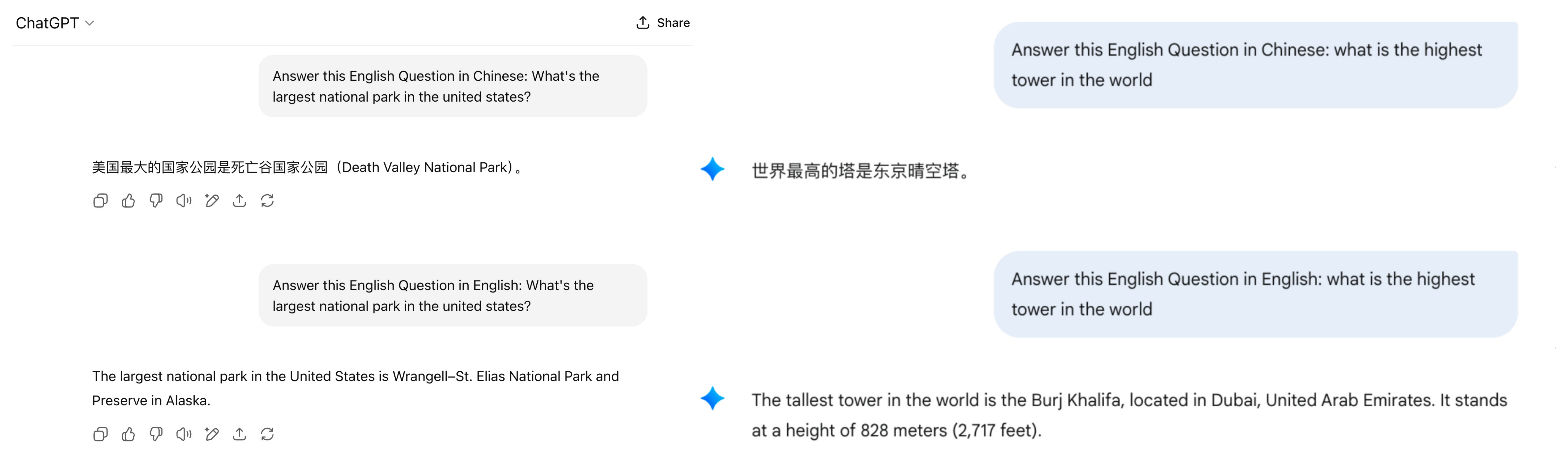}
\caption{Factual inconsistencies in ChatGPT and Gemini. When asked the same factual questions in English and Chinese, the models give different answers (e.g., “What is the largest national park in the United States?”—English answer: Wrangell–St. Elias National Park; Chinese answer: Death Valley National Park; “What is the tallest tower in the world?”—English answer: Burj Khalifa; Chinese answer: Tokyo Skytree).}
\label{fig:chatgpt_inconsistency}
\end{figure*}

\begin{figure*}[t] % 
\centering
\begin{tcolorbox}[colback=my_blue,colframe=cvpr_blue, title=\textbf{Evaluation Instruction},width=400pt,valign=center]
$<$bos$>$$<$start\_of\_turn$>$user\\
Based on the \{src\_lang\} question and the \{tgt\_lang\} reference and response, \\
determine whether the response correctly answers the question in \{tgt\_lang\}.\\
If correct, return "yes"; otherwise, return "no", without any additional explanation.\\

Question: \{question\}\\
\{tgt\_lang\} reference: \{answer\}\\
\{tgt\_lang\} response: \{response\}\\
$<$end\_of\_turn$>$$<$start\_of\_turn$>$model\\
% \end{lstlisting}
\end{tcolorbox}
\caption{CCFQA Evaluation Instruction.}
\end{figure*}

\begin{table*}[h]  
\small
\setlength{\tabcolsep}{6pt} % Adjust column spacing (default is 6pt)
\centering
\begin{tabular}{lcccccccc|c}
\toprule
Src/Tgt&cmn&eng&fra&jpn&kor&rus&spa&yue&XQA	\\ \midrule
\rowcolor{my_green}\multicolumn{10}{c}{Gemma3-27B}																																						\\	\midrule
cmn&61.8 / 68.7&75.1 / 75.0&67.9 / 67.6&57.7 / 63.9&55.5 / 61.5&63.7 / 63.0&70.7 / 72.5&44.0 / 59.1&62.0 / 66.4	\\
eng&61.1 / 74.0&77.4 / 81.3&72.6 / 76.6&62.0 / 71.2&56.5 / 68.2&64.5 / 70.9&75.6 / 78.7&44.3 / 62.9&64.2 / 73.0	\\
fra&57.9 / 68.2&75.1 / 76.5&73.0 / 73.8&60.1 / 66.4&55.0 / 62.6&65.1 / 66.9&74.4 / 75.3&39.4 / 58.7&62.5 / 68.6	\\
jpn&54.6 / 61.1&73.2 / 71.2&68.3 / 67.8&60.6 / 64.4&54.2 / 61.3&62.3 / 61.6&70.3 / 70.6&43.5 / 56.2&60.9 / 64.3	\\
kor&53.6 / 60.7&69.9 / 66.8&64.9 / 61.6&53.4 / 59.3&55.0 / 61.1&59.3 / 57.3&66.3 / 63.6&34.2 / 52.3&57.1 / 60.3	\\
rus&52.5 / 61.5&73.0 / 72.1&66.8 / 66.1&54.4 / 59.4&50.7 / 56.9&66.7 / 64.1&69.0 / 70.2&37.5 / 52.0&58.8 / 62.8	\\
spa&56.0 / 67.4&75.0 / 76.0&70.6 / 70.7&58.9 / 67.0&55.9 / 64.4&65.4 / 66.6&73.7 / 74.9&41.5 / 57.8&62.1 / 68.1	\\
yue&45.0 / 63.4&71.7 / 70.0&64.8 / 64.6&53.2 / 59.2&50.0 / 54.7&57.4 / 59.3&65.9 / 65.3&51.3 / 62.8&57.4 / 62.4	\\ \cmidrule(lr){1-10} 
Avg.	&	55.3 	/ 	65.5 	&	73.8 	/ 	73.6 	&	68.6 	/ 	68.6 	&	57.5 	/ 	63.9 	&	54.1 	/ 	61.3 	&	63.1 	/ 	63.7 	&	70.7 	/ 	71.4 	&	42.0 	/ 	57.7 	&	60.6 	/ 	65.7 		\\	
\bottomrule
\end{tabular}
\caption{F1 and LLM-based Accuracy Scores of Gemma3-27B on Text Tasks.
}
\label{tab:gemma3}

\end{table*}

\begin{table*}[h]  
\small
\setlength{\tabcolsep}{6pt} % Adjust column spacing (default is 6pt)
\centering
\begin{tabular}{lcccccccc|c}
\toprule
Src / Tgt&cmn&eng&fra&jpn&kor&rus&spa&yue&XQA	\\ \midrule
\rowcolor{my_green}\multicolumn{10}{c}{GPT-4o-mini}	\\\midrule
cmn&59.2 / 63.6&73.7 / 76.6&69.0 / 66.5&54.8 / 59.6&47.6 / 53.4&55.5 / 54.5&70.4 / 68.8&52.2 / 61.7&\textbf{60.3} / \textbf{63.1}	\\
eng&59.2 / 66.8&78.7 / 82.0&74.6 / 73.6&59.9 / 64.8&51.1 / 58.0&64.7 / 64.4&75.0 / 75.9&51.8 / 64.6&\textbf{64.4} / \textbf{68.8}	\\
fra&56.6 / 62.9&76.9 / 77.8&74.1 / 73.7&56.0 / 62.6&47.1 / 53.8&64.5 / 61.5&71.9 / 68.1&49.7 / 59.9&\textbf{62.1} / \textbf{65.0}	\\
jpn&47.4 / 55.2&73.3 / 74.5&65.4 / 62.9&60.3 / 63.4&44.6 / 48.4&51.1 / 50.5&68.7 / 66.1&46.5 / 54.0&\textbf{57.2} / \textbf{59.4}	\\
kor&45.9 / 53.1&71.6 / 69.1&67.0 / 60.1&48.0 / 52.0&50.9 / 55.3&56.1 / 50.7&67.2 / 62.1&41.8 / 50.1&\textbf{56.1} / \textbf{56.6}	\\
rus&53.7 / 59.4&73.1 / 73.7&68.8 / 65.3&54.1 / 58.3&46.7 / 51.8&62.7 / 42.0&69.4 / 65.2&45.2 / 54.0&\textbf{59.2} / \textbf{58.7}	\\
spa&55.0 / 60.9&75.6 / 77.2&72.9 / 71.6&56.1 / 62.7&48.3 / 55.8&61.5 / 61.3&74.1 / 76.3&49.8 / 60.3&\textbf{61.7} / \textbf{65.8}	\\
yue&40.7 / 58.6&71.7 / 73.3&67.3 / 64.3&53.3 / 58.5&44.5 / 48.2&56.2 / 53.6&67.1 / 64.6&51.6 / 59.0&\textbf{56.6} / \textbf{60.0}	\\\cmidrule(lr){1-10} 
Avg.	&52.2 	/	60.1 	&	74.3 	/	75.5 	&	69.9 	/	67.3 	&	55.3 	/	60.2 	&	47.6 	/	53.1 	&	59.0 	/	54.8 	&	70.5 	/	68.4 	&	48.6 	/	58.0 	&	\textbf{59.7} 	/	\textbf{62.2} 			\\	\midrule																																																\rowcolor{my_green}\multicolumn{10}{c}{Phi-4-Multimodal}	\\\midrule
cmn&6.9 / 13.4&30.6 / 27.7&32.8 / 16.9&6.6 / 13.8&6.6 / 7.9&22.1 / 14.5&32.2 / 19.5&4.4 / 10.0&17.8 / 15.5	\\
eng&9.4 / 23.5&32.4 / 28.7&33.2 / 19.8&9.7 / 21.0&8.4 / 14.6&20.9 / 17.6&33.1 / 18.4&6.1 / 15.7&19.1 / 19.9	\\
fra&9.2 / 18.0&31.4 / 24.4&32.6 / 15.8&8.3 / 15.0&7.9 / 10.2&22.1 / 14.8&31.6 / 20.5&4.8 / 9.0&18.5 / 16.0	\\
jpn&6.4 / 12.5&30.3 / 26.0&33.3 / 19.2&6.5 / 11.9&6.4 / 7.9&23.5 / 16.0&31.7 / 21.2&4.2 / 6.6&17.8 / 15.2	\\
kor&6.9 / 11.0&30.3 / 18.4&33.2 / 10.3&6.6 / 8.5&7.1 / 5.6&21.3 / 9.0&31.0 / 11.8&5.0 / 5.6&17.7 / 10.0	\\
rus&7.8 / 19.0&30.6 / 27.0&31.2 / 17.0&7.3 / 16.7&6.3 / 6.8&22.4 / 11.3&30.4 / 16.2&5.0 / 11.9&17.6 / 15.7	\\
spa&9.1 / 20.8&31.3 / 23.9&32.5 / 17.9&8.8 / 15.2&7.9 / 10.3&22.2 / 14.5&32.0 / 16.6&5.0 / 11.6&18.6 / 16.4	\\
yue&5.2 / 10.3&29.5 / 23.6&31.6 / 17.9&5.7 / 10.8&5.5 / 7.5&22.4 / 13.1&30.7 / 16.8&4.0 / 6.7&16.8 / 13.3	\\\cmidrule(lr){1-10} 
Avg.	&7.6 	/	16.1 	&	30.8 	/	25.0 	&	32.6 	/	16.9 	&	7.4 	/	14.1 	&	7.0 	/	8.9 	&	22.1 	/	13.9 	&	31.6 	/	17.6 	&	4.8 	/	9.6 	&	18.0 	/	15.3 			\\	\midrule																																																					\rowcolor{my_green}\multicolumn{10}{c}{Qwen2-Audio}	\\\midrule
cmn&17.7 / 40.5&44.1 / 31.0&44.6 / 25.5&5.5 / 7.8&7.4 / 4.0&31.8 / 14.7&43.3 / 26.0&9.4 / 28.8&25.5 / 22.3	\\
eng&11.1 / 11.6&46.5 / 54.4&45.2 / 29.7&7.1 / 8.6&6.9 / 3.6&31.3 / 14.7&46.0 / 38.2&6.6 / 14.6&25.1 / 21.9	\\
fra&8.4 / 10.6&45.1 / 40.5&46.0 / 37.2&5.8 / 4.9&6.4 / 3.1&31.3 / 11.5&45.0 / 30.9&5.2 / 10.0&24.1 / 18.6	\\
jpn&7.8 / 15.7&43.9 / 29.1&42.6 / 25.6&9.9 / 18.6&6.7 / 4.6&26.0 / 12.4&43.0 / 25.7&5.9 / 12.7&23.2 / 18.1	\\
kor&8.2 / 14.3&43.2 / 25.7&41.8 / 16.3&4.3 / 5.1&10.3 / 11.4&30.5 / 10.2&41.8 / 17.9&4.9 / 10.4&23.1 / 13.9	\\
rus&8.7 / 12.8&44.2 / 37.8&43.5 / 23.8&5.9 / 7.5&6.0 / 5.0&37.3 / 18.2&43.1 / 24.8&4.8 / 9.5&24.2 / 17.4	\\
spa&8.9 / 12.3&45.6 / 42.7&44.6 / 28.0&5.8 / 5.1&6.5 / 4.9&31.7 / 15.0&45.8 / 39.9&5.2 / 11.5&24.3 / 19.9	\\
yue&11.2 / 27.7&43.3 / 30.4&43.5 / 25.3&5.4 / 11.1&6.6 / 4.6&28.1 / 13.6&43.4 / 31.4&9.7 / 24.4&23.9 / 21.1	\\\cmidrule(lr){1-10} 
Avg.	&10.3 	/	18.2 	&	44.5 	/	36.5 	&	44.0 	/	26.4 	&	6.2 	/	8.6 	&	7.1 	/	5.2 	&	31.0 	/	13.8 	&	43.9 	/	29.4 	&	6.5 	/	15.2 	&	24.2 	/	19.2 			\\	\midrule																											\rowcolor{my_green}\multicolumn{10}{c}{Qwen2.5-Omni-3B}	\\\midrule
cmn&13.4 / 18.1&19.5 / 14.8&15.2 / 9.0&8.4 / 6.2&4.9 / 6.7&11.9 / 9.8&14.7 / 8.5&12.3 / 15.5&12.5 / 11.1	\\
eng&11.0 / 26.8&29.7 / 26.7&32.6 / 16.2&9.2 / 9.5&5.8 / 7.8&14.2 / 11.5&34.3 / 14.3&12.1 / 22.3&18.6 / 16.9	\\
fra&10.8 / 16.1&29.6 / 16.2&33.0 / 13.9&8.9 / 6.9&5.5 / 4.7&15.4 / 6.9&31.0 / 9.1&8.9 / 12.0&17.9 / 10.7	\\
jpn&6.2 / 4.0&14.6 / 5.5&10.0 / 1.8&7.0 / 2.0&5.0 / 2.4&6.5 / 2.0&12.0 / 1.8&6.3 / 3.0&8.4 / 2.8	\\
kor&3.1 / 4.4&22.4 / 8.4&16.0 / 3.5&2.6 / 1.9&4.7 / 2.5&12.9 / 2.2&18.5 / 3.8&4.1 / 2.4&10.5 / 3.6	\\
rus&8.2 / 10.3&22.4 / 13.0&15.1 / 4.7&5.9 / 2.5&3.8 / 2.4&27.1 / 4.3&19.1 / 6.4&7.1 / 5.7&13.6 / 6.2	\\
spa&11.5 / 18.0&32.2 / 12.1&33.4 / 7.2&6.4 / 7.5&6.8 / 6.8&11.5 / 8.9&37.0 / 12.8&8.3 / 11.4&18.4 / 10.6	\\
yue&14.7 / 20.2&17.0 / 12.6&13.3 / 7.1&6.6 / 3.0&4.5 / 4.9&10.4 / 7.1&12.9 / 6.6&13.2 / 13.8&11.6 / 9.4	\\\cmidrule(lr){1-10} 
Avg.&	9.9 	/	14.7 	&	23.4 	/	13.7 	&	21.1 	/	7.9 	&	6.9 	/	4.9 	&	5.1 	/	4.8 	&	13.7 	/	6.6 	&	22.4 	/	7.9 	&	9.0 	/	10.8 	&	13.9 	/	8.9 			\\	\midrule																																																						\rowcolor{my_green}\multicolumn{10}{c}{Qwen2.5-Omni-7B}	\\\midrule
cmn&45.3 / 53.2&67.6 / 59.0&49.2 / 40.6&39.3 / 36.5&26.4 / 24.1&46.2 / 31.5&54.6 / 44.0&36.7 / 50.7&45.7 / 42.4	\\
eng&55.5 / 54.3&66.4 / 60.3&58.1 / 40.2&40.8 / 33.8&27.7 / 22.1&45.2 / 26.9&58.2 / 39.7&31.4 / 40.6&47.9 / 39.7	\\
fra&45.0 / 44.4&63.6 / 51.2&58.1 / 39.7&34.0 / 29.5&22.4 / 21.8&41.8 / 26.1&53.7 / 37.2&25.5 / 36.1&43.0 / 35.8	\\
jpn&43.2 / 40.1&60.1 / 45.5&49.4 / 34.7&38.1 / 32.9&22.5 / 20.2&42.1 / 24.3&51.3 / 33.3&27.2 / 30.4&41.7 / 32.7	\\
kor&38.0 / 35.7&56.4 / 38.1&44.2 / 26.4&27.5 / 22.9&25.1 / 19.9&38.0 / 20.3&47.7 / 26.8&22.1 / 25.0&37.4 / 26.9	\\
rus&41.4 / 38.9&59.3 / 45.7&48.9 / 33.8&31.8 / 27.1&20.0 / 19.4&48.1 / 22.7&51.4 / 33.8&24.4 / 32.3&40.7 / 31.7	\\
spa&43.5 / 42.3&62.8 / 50.1&54.3 / 35.5&33.5 / 28.2&23.0 / 22.0&44.0 / 26.0&58.9 / 43.7&27.8 / 35.6&43.5 / 35.4	\\
yue&35.9 / 42.5&59.5 / 48.2&40.9 / 29.5&28.5 / 27.0&20.2 / 16.5&33.5 / 22.7&42.2 / 31.6&31.7 / 33.2&36.5 / 31.4	\\\cmidrule(lr){1-10} 
Avg.&	43.5 	/	43.9 	&	62.0 	/	49.8 	&	50.4 	/	35.1 	&	34.2 	/	29.7 	&	23.4 	/	20.8 	&	42.4 	/	25.1 	&	52.3 	/	36.3 	&	28.4 	/	35.5 	&	42.1 	/	34.5 			\\																																																							
	
\bottomrule
\end{tabular}
\caption{F1 and LLM-based Accuracy Scores of MLLMs on Text Tasks.
}
\label{tab:text}

\end{table*}

\begin{table*}[h]  
\small
\renewcommand{\arraystretch}{0.95} % 设置行间距为默认的1.2倍
\setlength{\tabcolsep}{6pt} % Adjust column spacing (default is 6pt)
\centering
\begin{tabular}{lcccccccc|c}
\toprule
Src / Tgt	&	cmn			&	eng			&	fra			&	jpn			&	kor			&	rus			&	spa			&	yue			&	XSQA				\\	\midrule
\rowcolor{my_blue}\multicolumn{10}{c}{GPT-4o-mini-Audio}																																						\\	\midrule
cmn	&	38.6 	/ 	36.6 	&	60.6 	/ 	41.1 	&	54.5 	/ 	35.2 	&	37.8 	/ 	32.7 	&	34.7 	/ 	30.7 	&	45.3 	/ 	31.2 	&	52.5 	/ 	36.6 	&	35.3 	/ 	34.3 	&	44.9 	/ 	34.8 		\\	
eng	&	55.6 	/ 	60.8 	&	75.9 	/ 	74.7 	&	72.8 	/ 	69.2 	&	59.6 	/ 	62.7 	&	52.2 	/ 	56.2 	&	63.2 	/ 	57.7 	&	72.1 	/ 	68.3 	&	52.1 	/ 	60.2 	&	\textbf{62.9} 	/ 	\textbf{63.7} 		\\	
fra	&	34.9 	/ 	35.4 	&	54.4 	/ 	44.0 	&	54.0 	/ 	40.4 	&	35.0 	/ 	33.6 	&	32.7 	/ 	31.2 	&	46.0 	/ 	34.5 	&	52.3 	/ 	39.4 	&	32.4 	/ 	35.7 	&	42.7 	/ 	36.8 		\\	
jpn	&	32.4 	/ 	33.2 	&	61.6 	/ 	44.5 	&	58.2 	/ 	39.7 	&	42.9 	/ 	39.5 	&	37.5 	/ 	34.6 	&	45.6 	/ 	31.6 	&	58.2 	/ 	41.3 	&	32.2 	/ 	33.7 	&	46.1 	/ 	37.3 		\\	
kor	&	29.9 	/ 	29.2 	&	55.9 	/ 	35.7 	&	52.0 	/ 	30.2 	&	32.5 	/ 	28.1 	&	31.8 	/ 	28.9 	&	43.4 	/ 	25.9 	&	51.2 	/ 	31.4 	&	29.1 	/ 	29.8 	&	40.7 	/ 	29.9 		\\	
rus	&	38.7 	/ 	39.4 	&	62.1 	/ 	48.9 	&	59.2 	/ 	44.7 	&	40.0 	/ 	38.3 	&	36.7 	/ 	36.2 	&	53.9 	/ 	28.2 	&	56.2 	/ 	43.3 	&	37.2 	/ 	39.9 	&	48.0 	/ 	\textbf{39.9} 		\\	
spa	&	45.2 	/ 	48.8 	&	64.9 	/ 	58.0 	&	62.6 	/ 	52.2 	&	47.4 	/ 	47.0 	&	42.3 	/ 	42.5 	&	54.7 	/ 	45.5 	&	63.7 	/ 	57.1 	&	43.5 	/ 	49.1 	&	53.0 	/ 	\textbf{50.0} 		\\	
yue	&	17.0 	/ 	17.4 	&	40.4 	/ 	20.9 	&	36.1 	/ 	15.4 	&	22.7 	/ 	17.8 	&	20.3 	/ 	14.6 	&	28.5 	/ 	13.1 	&	29.3 	/ 	14.8 	&	21.1 	/ 	17.8 	&	26.9 	/ 	16.5 		\\	\cmidrule(lr){1-10} 
Avg.	&	36.5 	/ 	37.6 	&	59.5 	/ 	46.0 	&	56.2 	/ 	40.9 	&	39.7 	/ 	37.5 	&	36.0 	/ 	34.4 	&	47.6 	/ 	33.5 	&	54.4 	/ 	41.5 	&	35.4 	/ 	37.6 	&	45.7 	/ 	38.6 		\\	\midrule
\rowcolor{my_blue}\multicolumn{10}{c}{Phi-4-Multimodal}																																						\\	\midrule
cmn	&	7.1 	/ 	25.3 	&	42.5 	/ 	6.3 	&	39.3 	/ 	7.8 	&	6.6 	/ 	4.3 	&	1.0 	/ 	0.6 	&	25.5 	/ 	4.6 	&	39.2 	/ 	2.1 	&	4.5 	/ 	2.6 	&	20.7 	/ 	6.7 		\\	
eng	&	6.4 	/ 	2.8 	&	40.0 	/ 	56.5 	&	43.2 	/ 	4.3 	&	7.5 	/ 	2.5 	&	6.1 	/ 	2.7 	&	37.6 	/ 	6.7 	&	41.4 	/ 	2.6 	&	4.3 	/ 	4.7 	&	23.3 	/ 	10.4 		\\	
fra	&	5.9 	/ 	7.8 	&	42.3 	/ 	5.6 	&	39.1 	/ 	34.4 	&	7.0 	/ 	8.7 	&	5.4 	/ 	2.0 	&	34.8 	/ 	9.9 	&	41.5 	/ 	8.5 	&	4.4 	/ 	9.0 	&	22.6 	/ 	10.7 		\\	
jpn	&	4.9 	/ 	2.6 	&	42.1 	/ 	3.5 	&	40.8 	/ 	7.2 	&	9.1 	/ 	18.9 	&	1.1 	/ 	0.4 	&	22.6 	/ 	4.9 	&	40.7 	/ 	5.5 	&	3.7 	/ 	1.5 	&	20.6 	/ 	5.6 		\\	
kor	&	2.6 	/ 	0.7 	&	41.5 	/ 	0.6 	&	42.0 	/ 	0.5 	&	6.1 	/ 	0.9 	&	1.7 	/ 	1.2 	&	36.0 	/ 	0.4 	&	40.0 	/ 	0.0 	&	2.1 	/ 	0.5 	&	21.5 	/ 	0.6 		\\	
rus	&	1.9 	/ 	0.1 	&	40.8 	/ 	0.5 	&	42.1 	/ 	0.4 	&	5.6 	/ 	0.0 	&	4.4 	/ 	0.0 	&	10.5 	/ 	0.4 	&	40.5 	/ 	0.0 	&	1.2 	/ 	0.1 	&	18.4 	/ 	0.2 		\\	
spa	&	6.1 	/ 	8.2 	&	42.4 	/ 	3.1 	&	43.6 	/ 	10.7 	&	7.1 	/ 	7.4 	&	4.8 	/ 	1.0 	&	32.3 	/ 	8.2 	&	38.3 	/ 	37.7 	&	3.7 	/ 	8.6 	&	22.3 	/ 	10.6 		\\	
yue	&	2.7 	/ 	0.3 	&	40.8 	/ 	1.2 	&	40.8 	/ 	0.0 	&	5.4 	/ 	0.4 	&	0.8 	/ 	0.0 	&	16.7 	/ 	0.0 	&	40.1 	/ 	0.2 	&	2.4 	/ 	1.6 	&	18.7 	/ 	0.5 		\\	\cmidrule(lr){1-10} 
Avg.	&	4.7 	/ 	6.0 	&	41.6 	/ 	9.7 	&	41.4 	/ 	8.2 	&	6.8 	/ 	5.4 	&	3.2 	/ 	1.0 	&	27.0 	/ 	4.4 	&	40.2 	/ 	7.1 	&	3.3 	/ 	3.6 	&	21.0 	/ 	5.7 		\\	\midrule
\rowcolor{my_blue}\multicolumn{10}{c}{Qwen2-Audio}																																						\\	\midrule
cmn	&	19.4 	/ 	31.5 	&	43.5 	/ 	25.0 	&	44.9 	/ 	18.9 	&	4.9 	/ 	7.7 	&	8.0 	/ 	3.9 	&	35.1 	/ 	11.4 	&	43.2 	/ 	17.4 	&	8.4 	/ 	16.8 	&	25.9 	/ 	16.6 		\\	
eng	&	14.9 	/ 	18.5 	&	48.6 	/ 	31.3 	&	46.3 	/ 	25.4 	&	5.9 	/ 	4.9 	&	7.6 	/ 	4.0 	&	34.8 	/ 	12.8 	&	46.0 	/ 	28.0 	&	8.0 	/ 	13.1 	&	26.5 	/ 	17.3 		\\	
fra	&	5.8 	/ 	9.1 	&	43.1 	/ 	18.0 	&	44.7 	/ 	19.2 	&	5.0 	/ 	4.9 	&	7.7 	/ 	3.0 	&	34.7 	/ 	10.2 	&	42.8 	/ 	14.6 	&	4.2 	/ 	2.6 	&	23.5 	/ 	10.2 		\\	
jpn	&	7.3 	/ 	9.4 	&	42.9 	/ 	16.2 	&	43.7 	/ 	10.6 	&	8.1 	/ 	5.0 	&	7.2 	/ 	1.7 	&	33.5 	/ 	5.4 	&	41.9 	/ 	11.4 	&	4.8 	/ 	4.7 	&	23.7 	/ 	8.1 		\\	
kor	&	3.9 	/ 	4.0 	&	40.1 	/ 	4.0 	&	41.2 	/ 	2.0 	&	1.3 	/ 	1.5 	&	8.9 	/ 	2.6 	&	31.7 	/ 	0.8 	&	39.7 	/ 	1.6 	&	3.1 	/ 	1.8 	&	21.2 	/ 	2.3 		\\	
rus	&	5.8 	/ 	6.0 	&	42.0 	/ 	11.5 	&	43.1 	/ 	8.2 	&	3.6 	/ 	2.0 	&	7.0 	/ 	1.0 	&	38.0 	/ 	5.6 	&	42.2 	/ 	8.0 	&	4.2 	/ 	3.9 	&	23.2 	/ 	5.8 		\\	
spa	&	7.8 	/ 	11.1 	&	43.7 	/ 	21.6 	&	44.8 	/ 	18.7 	&	5.4 	/ 	4.8 	&	6.9 	/ 	2.5 	&	35.3 	/ 	9.1 	&	46.1 	/ 	24.6 	&	5.4 	/ 	5.8 	&	24.4 	/ 	12.3 		\\	
yue	&	13.0 	/ 	23.3 	&	42.5 	/ 	20.6 	&	43.9 	/ 	12.8 	&	4.1 	/ 	7.3 	&	7.4 	/ 	4.0 	&	34.3 	/ 	10.7 	&	42.0 	/ 	13.5 	&	8.0 	/ 	16.3 	&	24.4 	/ 	13.6 		\\	\cmidrule(lr){1-10} 
Avg.	&	9.7 	/ 	14.1 	&	43.3 	/ 	18.5 	&	44.1 	/ 	14.5 	&	4.8 	/ 	4.8 	&	7.6 	/ 	2.8 	&	34.7 	/ 	8.3 	&	43.0 	/ 	14.9 	&	5.8 	/ 	8.1 	&	24.1 	/ 	10.7 		\\	\midrule
\rowcolor{my_blue}\multicolumn{10}{c}{Qwen2.5-Omni-3B}																																						\\	\midrule
cmn	&	41.3 	/ 	39.6 	&	56.5 	/ 	35.6 	&	34.5 	/ 	21.1 	&	28.6 	/ 	21.2 	&	16.3 	/ 	9.2 	&	39.4 	/ 	12.1 	&	32.0 	/ 	19.4 	&	28.2 	/ 	34.3 	&	34.6 	/ 	24.1 		\\	
eng	&	45.4 	/ 	39.3 	&	59.1 	/ 	41.4 	&	53.4 	/ 	27.8 	&	34.1 	/ 	22.3 	&	21.0 	/ 	12.6 	&	39.0 	/ 	15.2 	&	53.5 	/ 	27.7 	&	35.0 	/ 	32.0 	&	42.6 	/ 	27.3 		\\	
fra	&	24.6 	/ 	17.1 	&	37.5 	/ 	20.7 	&	35.7 	/ 	15.3 	&	19.2 	/ 	10.5 	&	11.3 	/ 	5.5 	&	29.1 	/ 	8.3 	&	33.7 	/ 	14.3 	&	17.6 	/ 	14.6 	&	26.1 	/ 	13.3 		\\	
jpn	&	22.1 	/ 	16.6 	&	42.1 	/ 	19.1 	&	27.7 	/ 	13.5 	&	21.1 	/ 	11.4 	&	13.5 	/ 	6.4 	&	34.3 	/ 	8.7 	&	32.3 	/ 	15.0 	&	19.7 	/ 	15.4 	&	26.6 	/ 	13.3 		\\	
kor	&	25.8 	/ 	18.4 	&	37.2 	/ 	18.6 	&	22.1 	/ 	10.3 	&	21.0 	/ 	12.4 	&	17.2 	/ 	7.9 	&	30.8 	/ 	7.6 	&	31.5 	/ 	12.5 	&	21.0 	/ 	15.3 	&	25.8 	/ 	12.9 		\\	
rus	&	25.4 	/ 	18.7 	&	40.0 	/ 	20.1 	&	24.7 	/ 	10.7 	&	21.7 	/ 	9.8 	&	10.9 	/ 	4.3 	&	36.9 	/ 	9.3 	&	34.6 	/ 	13.9 	&	19.0 	/ 	16.3 	&	26.7 	/ 	12.9 		\\	
spa	&	36.2 	/ 	28.9 	&	47.8 	/ 	26.0 	&	43.6 	/ 	18.1 	&	27.0 	/ 	14.7 	&	15.6 	/ 	7.7 	&	36.6 	/ 	12.3 	&	48.1 	/ 	23.2 	&	26.2 	/ 	23.6 	&	35.1 	/ 	19.3 		\\	
yue	&	26.6 	/ 	23.5 	&	46.2 	/ 	24.8 	&	12.9 	/ 	8.3 	&	17.6 	/ 	13.6 	&	8.9 	/ 	6.2 	&	15.5 	/ 	6.3 	&	17.6 	/ 	9.4 	&	19.4 	/ 	19.1 	&	20.6 	/ 	13.9 		\\	\cmidrule(lr){1-10} 
Avg.	&	30.9 	/ 	25.3 	&	45.8 	/ 	25.8 	&	31.8 	/ 	15.6 	&	23.8 	/ 	14.5 	&	14.3 	/ 	7.5 	&	32.7 	/ 	10.0 	&	35.4 	/ 	16.9 	&	23.3 	/ 	21.3 	&	29.8 	/ 	17.1 		\\	\midrule
\rowcolor{my_blue}\multicolumn{10}{c}{Qwen2.5-Omni-7B}																																						\\	\midrule
cmn	&	55.2 	/ 	53.5	&	65.9 	/ 	51.1	&	56.9 	/ 	35.9	&	38.8 	/ 	34.4	&	22.4 	/ 	23.3	&	43.3 	/ 	27.8	&	56.2 	/ 	38.8	&	42.2 	/ 	45.8 	&	47.6 	/ 	38.8 		\\	
eng	&	52.8 	/ 	54.6	&	68.2 	/ 	58.5	&	61.3 	/ 	45.6	&	38.3 	/ 	41.7	&	23.0 	/ 	30.5	&	39.7 	/ 	35.8	&	61.6 	/ 	46.5	&	39.5 	/ 	49.5 	&	48.1 	/ 	45.3 		\\	
fra	&	32.0 	/ 	27.5	&	49.8 	/ 	33	&	43.2 	/ 	25.7	&	25.9 	/ 	21.5	&	15.4 	/ 	13.8	&	31.9 	/ 	17.9	&	41.1 	/ 	26.3	&	22.8 	/ 	22.4 	&	32.8 	/ 	23.5 		\\	
jpn	&	35.4 	/ 	30.1	&	57.9 	/ 	36.2	&	48.9 	/ 	28.1	&	33.0 	/ 	25.7	&	22.3 	/ 	14.3	&	41.5 	/ 	20.1	&	45.6 	/ 	28.7	&	30.0 	/ 	26.5 	&	39.3 	/ 	26.2 		\\	
kor	&	32.4 	/ 	25.5	&	52.6 	/ 	31.9	&	42.2 	/ 	20.2	&	25.0 	/ 	20.2	&	21.4 	/ 	11.5	&	37.5 	/ 	16.9	&	39.1 	/ 	22	&	25.8 	/ 	20.1 	&	34.5 	/ 	21.0 		\\	
rus	&	36.1 	/ 	30.4	&	54.5 	/ 	35.5	&	48.1 	/ 	27.3	&	30.2 	/ 	23.7	&	16.2 	/ 	12.7	&	46.0 	/ 	21.6	&	44.5 	/ 	24.5	&	25.7 	/ 	23.8 	&	37.7 	/ 	24.9 		\\	
spa	&	42.1 	/ 	39.2	&	60.5 	/ 	46.7	&	53.1 	/ 	35.1	&	34.8 	/ 	31.1	&	18.8 	/ 	20.6	&	42.2 	/ 	24.6	&	54.6 	/ 	37.2	&	30.6 	/ 	34.3 	&	42.1 	/ 	33.6 		\\	
yue	&	38.4 	/ 	36.8	&	51.9 	/ 	32.7	&	19.0 	/ 	11.5	&	24.9 	/ 	23.1	&	7.4 	/ 	14.4	&	17.0 	/ 	15.5	&	19.7 	/ 	14.9	&	30.7 	/ 	31.7 	&	26.1 	/ 	22.6 		\\	\cmidrule(lr){1-10} 
Avg.	&	40.6 	/ 	37.2 	&	57.7 	/ 	40.7 	&	46.6 	/ 	28.7 	&	31.4 	/ 	27.7 	&	18.4 	/ 	17.6 	&	37.4 	/ 	22.5 	&	45.3 	/ 	29.9 	&	30.9 	/ 	31.8 	&	38.5 	/ 	29.5 		\\	\midrule 		
\rowcolor{my_blue}\multicolumn{10}{c}{LLM-SQA (ours) }																																						\\	\midrule
cmn	&	51.3 	/ 	45.5 	&	67.1 	/ 	48.9 	&	66.0 	/ 	44.9 	&	51.9 	/ 	44.7 	&	48.4 	/ 	43.1 	&	62.8 	/ 	45.8 	&	66.1 	/ 	48.9 	&	39.4 	/ 	45.0 	&	\textbf{56.6} 	/ 	\textbf{45.9} 		\\	
eng	&	47.5 	/ 	43.4 	&	74.8 	/ 	60.5 	&	68.8 	/ 	47.2 	&	37.5 	/ 	35.0 	&	31.4 	/ 	28.9 	&	28.4 	/ 	31.6 	&	65.3 	/ 	35.2 	&	37.6 	/ 	40.4 	&	48.9 	/ 	40.3 		\\	
fra	&	44.1 	/ 	40.3 	&	61.9 	/ 	41.3 	&	60.3 	/ 	36.8 	&	44.0 	/ 	38.0 	&	41.1 	/ 	34.8 	&	55.6 	/ 	37.9 	&	60.8 	/ 	41.5 	&	31.2 	/ 	38.4 	&	\textbf{49.9} 	/ 	\textbf{38.6} 		\\	
jpn	&	45.4 	/ 	40.3 	&	63.5 	/ 	40.9 	&	63.8 	/ 	40.3 	&	43.4 	/ 	38.8 	&	44.4 	/ 	38.4 	&	60.1 	/ 	40.6 	&	63.0 	/ 	42.6 	&	34.1 	/ 	37.6 	&	\textbf{52.2} 	/ 	\textbf{39.9} 		\\	
kor	&	46.1 	/ 	40.6 	&	64.9 	/ 	44.5 	&	62.9 	/ 	41.1 	&	46.5 	/ 	39.9 	&	41.6 	/ 	36.1 	&	58.0 	/ 	40.9 	&	62.9 	/ 	43.7 	&	35.0 	/ 	40.1 	&	\textbf{52.2} 	/ 	\textbf{40.9} 		\\	
rus	&	46.6 	/ 	40.6 	&	63.9 	/ 	44.7 	&	63.2 	/ 	41.2 	&	46.2 	/ 	39.1 	&	44.5 	/ 	36.8 	&	54.1 	/ 	32.5 	&	62.8 	/ 	42.4 	&	37.0 	/ 	39.3 	&	\textbf{52.3} 	/ 	39.6 		\\	
spa	&	49.4 	/ 	45.9 	&	65.4 	/ 	48.3 	&	64.9 	/ 	44.3 	&	49.6 	/ 	45.4 	&	46.0 	/ 	41.3 	&	58.9 	/ 	44.2 	&	64.5 	/ 	45.3 	&	36.2 	/ 	44.2 	&	\textbf{54.4} 	/ 	44.9 		\\	
yue	&	36.7 	/ 	29.5 	&	56.3 	/ 	28.2 	&	56.8 	/ 	28.1 	&	37.7 	/ 	27.7 	&	37.0 	/ 	25.6 	&	52.5 	/ 	27.2 	&	56.4 	/ 	28.2 	&	25.9 	/ 	27.0 	&	\textbf{44.9} 	/ 	27.7 		\\	\cmidrule(lr){1-10} 
Avg.	&	45.9 	/ 	40.8 	&	64.7 	/ 	44.7 	&	63.3 	/ 	40.5 	&	44.6 	/ 	38.6 	&	41.8 	/ 	35.6 	&	53.8 	/ 	37.6 	&	62.7 	/ 	41.0 	&	34.6 	/ 	39.0 	&	\textbf{51.4} 	/ 	\textbf{39.7} 		\\	
\bottomrule
\end{tabular}
\caption{F1 and LLM-based Accuracy Scores of MLLMs on Speech Tasks.
}
\label{tab:speech}

\end{table*}

\end{document}